\documentclass{article}

\usepackage[nonatbib,preprint]{neurips_2025}
\usepackage{ulem}
\usepackage{subcaption} 

\usepackage[utf8]{inputenc} 
\usepackage[T1]{fontenc}    
\usepackage{hyperref}       
\usepackage{url}            
\usepackage{booktabs}       
\usepackage{amsfonts}       
\usepackage{nicefrac}       
\usepackage{microtype}      
\usepackage{xcolor}         
\usepackage{amsmath}
\usepackage{todonotes}
\usepackage{multirow} 
\usepackage{titlesec}
\usepackage{tcolorbox}
\usepackage{booktabs}
\usepackage{tabularx}
\usepackage{subcaption}
\usepackage{cleveref}
\usepackage[table]{xcolor}  
\usepackage{enumitem}
\usepackage[numbers]{natbib}

\newtheorem{thm}{Theorem}[section]

\title{Pluralistic Off-policy Evaluation and Alignment}

\author{
Chengkai Huang$^{1,5}\thanks{These authors contributed equally to this work.}$,
Junda Wu$^{2*}$, Zhouhang Xie$^2$, Yu Xia$^2$, Rui Wang$^3$, \\
\textbf{Tong Yu$^3$, Subrata Mitra$^3$, Julian McAuley$^2$, Lina Yao$^{1,4}$}
 \\
$^1$The University of New South Wales \quad 
$^2$UC San Diego \quad 
$^3$Adobe Research \\
$^4$CSIRO's Data61  \quad 
$^5$Macquarie University \\
\texttt{chengkai.huang1@unsw.edu.au} \\
\texttt{\{juw069,zhx022,yux078,jmcauley\}@ucsd.edu} \\
\texttt{\{tyu,sumitra\}@adobe.com}  \quad
\texttt{lina.yao@data61.csiro.au} \\
}

\begin{document}

\maketitle

\begin{abstract}




Personalized preference alignment for LLMs with diverse human preferences requires evaluation and alignment methods that capture pluralism. 
Most existing preference alignment datasets are logged under policies that differ substantially from the evaluated LLMs, and existing off‐policy estimators focus solely on overall utility while ignoring preference pluralism. Extending Off-Policy Evaluation (OPE) to pluralistic preference alignment, therefore, remains an open question.
Thus, we propose the Pluralistic Off-Policy Evaluation (\texttt{POPE}), the first framework for offline pluralistic preference evaluation and alignment in LLMs.
\texttt{POPE} includes a unified reward function that combines (1) a collaborative utility component derived from human preference signals (e.g., upvotes or relevance scores) and (2) a diversity component inspired by entropy-based coverage measures, together reflecting pluralistic alignment. Furthermore, to estimate this reward from logged interactions, we derive decomposable inverse propensity scoring (IPS) estimators that separately evaluate relevance and diversity.
Theoretically, we prove that our decomposed IPS estimators establish a lower bound on their variance. 
With the off-policy evaluated value function, we can directly enable off-policy optimization to further enhance pluralistic alignment. 
Empirical results demonstrate that \texttt{POPE} efficiently enhances pluralistic response generation and maintains the models’ general capabilities on downstream tasks.



\end{abstract}

\section{Introduction}






Preference-aligned language models can generate responses aligned with general human preferences, but recent works show they failed to capture diverse preferences among various acceptable responses, motivating pluralistic alignment~\cite{Sorensen2024RoadmapPluralistic,chen2024palpluralisticalignmentframework,feng2024modular} for large language models.
Ideally, a pluralistically aligned language model should be able to prioritize helpful responses while covering a diverse range of potentially acceptable responses, even if these responses do not represent majority preferences.
Such property is beneficial for various use cases, such as catering to diverse and potentially conflicting user preferences~\cite{Joel2023personalizedsoup} and modeling varied and diverging responses to provide a comprehensive view for the corresponding input~\cite{heyueya2024psychometricalignmentcapturinghuman,agentsurvey,fmsurvey}.


A key challenge in achieving pluralistic alignment is developing a language model evaluation
that balances two key properties: \textit{collaborative utility},
the overall value of its responses judged by the whole user population \cite{zheng2015unexpectedness}, and \textit{diversity}, the range of valid response the model is able to cover, similar to trade-off between high-reward and long-tail labels in adjacent domains fair machine learning and recommender system~\cite{ZHE2023DLforSerendipity, Simon2024FairnessML,bias2}.
Unfortunately, prior work in optimizing and evaluating pluralistic responses in language models typically only focuses on replicating the target preference distribution~\cite{Shibani2023whose, durmus2024towards, chen2024palpluralisticalignmentframework,zhao2023group}, which overlooks the \textit{collaborative utility} aspect.
On the other hand, as shown in prior work~\cite{Sorensen2024RoadmapPluralistic} and later in our experiments (\Cref{tab:alpaca_gpt4}), pure utility-based alignment methods such as reward-model based methods \cite{ji2025survey} and DPO \cite{rafailov2023direct} will bias the model towards a narrow set of high-reward responses \cite{fu2025reward,chen2024odin}.
To this end, training and evaluating models that can cater to diverse user preferences while producing globally aligned responses remains an open problem. 
Meanwhile, current preference alignment datasets are typically collected using responses sampled from a different policy to the target model being evaluated/optimized, an ideal framework should work well with a logged dataset of responses and preferences collected from arbitrary sources. 
To this end, while there exist various attempts to develop off-policy evaluation for language model alignment \cite{wu2024ocean,bhargava2024off}.
Prior art typically considers only the global utility, leaving off-policy evaluation for pluralistic preferences for language models an open problem.


To address these challenges, we define a unified reward function combining: \textbf{Collaborative Utility reward} from human preference feedback, such as upvotes or relevancy scores, and \textbf{Diversity reward} inspired by serendipity and entropy-based measures of token-level coverage across the response space. This enables a principled value function that captures pluralistic preference alignment.
We propose a set of estimators that decompose the OPE objective into relevance and diversity components. For collaborative utility, we derive a standard inverse propensity score (IPS) estimator that reweights human-annotated preferences. For diversity, we derive a soft-entropy-based estimator inspired by Soft Q-Learning~\cite{haarnoja2017reinforcement}, enabling measurement of response space coverage from offline logs.

To be specific, we propose a new framework for evaluating and optimizing language models that explicitly accounts for both response \textit{utility} and \textit{diversity}, termed \textbf{Pluralistic Off-Policy Evaluation} (\texttt{POPE}). 
We prove that our policy estimator provides a lower bound for the variance. To enable direct optimization of LLM policies, we leverage the proposed \texttt{POPE} policy evaluation approach for LLM fine-tuning by directly maximizing estimated policy values through gradient descent. Then we empirically evaluate the optimized LLM policy on various open-ended response generation tasks, and demonstrate the effectiveness of the proposed policy optimization method, without affecting LLMs’ generalizability or generation quality. 
Our main contributions are as follows:
\begin{itemize}
    \item We introduce \texttt{POPE}, the first framework to apply off‐policy evaluation for measuring pluralistic alignment in LLMs.      
    \item We formalize a unified value function combining collaborative utility and diversity rewards to capture pluralistic preference alignment. Meanwhile, we propose decomposable IPS-based estimators that robustly evaluate utility and diversity from human-logged data.
    \item Within the evaluation framework, we propose a differentiable policy-optimization scheme that aligns LLMs on a unified reward combining utility and diversity.
    \item We demonstrate the effectiveness of our method on open-ended response generation datasets, showing improved alignment with pluralistic human responses without affecting LLMs’ generalizability.
\end{itemize}

\section{Related Work}

\textbf{Pluralistic Alignment.} While general alignment methods for learning from human feedback~\cite {rafailov2023direct, ouyang2022traininglanguagemodelsfollow, stiennon2022learningsummarizehumanfeedback} aim to produce language models that cater to the overall preferences of target users, human preferences are naturally diverse~\cite{jang2023personalizedsoupspersonalizedlarge, zhang2024divergingpreferencesannotatorsdisagree}.
Recently, there have been emerging efforts to capture such variability in language model optimization, building language models that cater to the preferences of groups of users~\cite{kumar2024compo, feng2024modular}, such alignment process is often termed as pluralistic alignment~\cite{Sorensen2024RoadmapPluralistic, xie2025surveypersonalizedpluralisticpreference}.
To this end, existing work most frequently build steerable models that can adapt their behavior conditioned on user specific inputs~\cite{lee2024aligning, wang2024learningpersonalizedalignmentevaluating, kumar2024compo, poddar2024personalizing, chen2024pal, balepur2025whose}, which are not always available.
Meanwhile, other works that attempt to incorporate pluralistic preferences into alignment~\cite{yao2025no, li2024aligningcrowdfeedbackdistributional, siththaranjan2024distributional}
Typically focus on an online setting.
To this end, to the best of our knowledge, we are the first work that tackles offline evaluation and optimization for pluralistic alignment.

\textbf{Offline Policy Evaluation.} 
Recent advances have adapted offline policy evaluation from reinforcement learning to language models.
Recent studies \citep{bhargava2024off,gao2023off} re-examined importance-sampling, doubly robust, and fitted-Q estimators for reward models trained on logged human feedback, showing that they remain unbiased and have competitive variance when the logging policy is close to the target.
Building on that line of work, OCEAN \citep{wu2024ocean} introduces knowledge-graph-grounded inverse propensity scores to assess chain-of-thought traces.
Theoretical studies analyze when model-based estimators tighten confidence intervals in RLHF \citep{li2024policy}.
Offline preference optimization methods \citep{ji2024self,zhou2024wpo,yin2024self} have also been developed which rely on these OPE guarantees to select gradient targets without new human queries, underscoring the need for accurate value estimation during alignment.

\textbf{Serendipity and Fairness for User Preferences and Diversity.}
Beyond pure accuracy, modern recommender systems increasingly aim to support pluralistic user preferences and result diversity by explicitly optimizing for both serendipity (unexpected yet relevant items) and fairness (equitable treatment of users or items) \cite{duricic2023beyond,bias2}.
Diversity has been proposed to demonstrate
that multi‐objective recommendation can mitigate filter‐bubble effects and promote under‐represented viewpoints \cite{knees2023recommender}. Definitions, metrics, and algorithms for serendipity have been reviewed to emphasize the balance between novelty and relevance \cite{ziarani2021serendipity}. The intrinsic links between fairness and diversity have been highlighted through unified frameworks that treat user‐level fairness as a form of diversity \cite{zhao2023fairness}. Algorithmic models have been developed to capture each user’s ideal trade‐off between diversity and serendipity, enabling adaptive recommendation strategies \cite{yin2024diversity}. In news recommendation, approaches to achieve media pluralism range from platform‐driven nudges for diversity to user‐controlled settings, ensuring exposure to varied perspectives while preserving user autonomy \cite{vermeulen2022nudge}.

\section{Problem Formulation}

We study the following setting: A conversational policy (\textit{e.g.}, an LLM or a human user) interacts with a human user by answering a query $x_t\in\mathcal{X}$.
For each query, the LLM generates a set of $K$ responses $S_t=\{a_{t,i}\}_{i=1}^K$, catering to the user's potentially \textit{diverse} preference.
This policy \textit{aligns with users' pluralistic preferences} \cite{Joel2023personalizedsoup,Sorensen2024RoadmapPluralistic} in conversational settings to reflect \textit{the wisdom of crowds} \cite{welinder2010multidimensional,huang1,huang2,zhiqiang}.


Assume there are $L=|\mathcal{S}|$ potential responses, where each action $a\in \mathcal{S}$ is sampled from it.
Unlike \cite{bhargava2024off}, we do not restrict slate size; our framework accommodates any generated responses that are less than or equal to all potential responses $K \le L$, though in practice, smaller $K$ reduces human annotation burden. By presenting multiple options, the policy seeks to capture users’ pluralistic preferences—the “wisdom of crowds” in conversational feedback \cite{welinder2010multidimensional,Joel2023personalizedsoup,Sorensen2024RoadmapPluralistic}.
We collect human feedback as a ranking (permutation) over the \(K\) presented responses. When \(K=2\), this reduces to pairwise comparisons as in the Bradley–Terry–Luce model \cite{bradley1952rank}; for \(K>2\), it corresponds to full-slate rankings via the Plackett–Luce model \cite{plackett1975analysis,debreu1960individual}. We will next define both the policy’s generation process and its alignment with pluralistic human rankings.

\paragraph{Policy.} Open-ended responses can vary greatly in length and structure, making raw token-level probabilities unstable for comparing full responses. To address this, we define two complementary views of our policy $\pi$:  

\textbf{(i) Token-level action.} We define a length-normalized sequence probability by,
\begin{align}
\pi(a\mid x) = \exp\Bigl(\frac{1}{|a|}\sum_{h=1}^{|a|}\log p_\theta(a_h\mid a_{<h},x)\Bigr).
\end{align}
The probability $\pi(a\mid x)$ is obtained not via separate token‐level log‐likelihood \cite{bhargava2024off,wu2024ocean}, but by exponentiating the average token log‐likelihood over the entire response,
Where $|a|$ is the total number of tokens in the response $a$, and $a_h$ denotes the $h$-th token of $a$. This choice reflects our focus on pluralistic alignment in open‐ended tasks: human logged data exhibit diverse, variable‐length responses whose token‐level scores are not directly comparable, so using the average log‐likelihood yields a length‐normalized, stable measure of human preference.

\textbf{(ii) Response-level action.}  $S$ denotes set of $K$ responses $S=\{a_{i}\}_{i=1}^K$ is sampled from the policy $\pi(\cdot\mid x)$  with the probability,
\begin{equation}
    \pi(S\mid x) = \sum_{i=1}^k  \frac{\pi(a_i\mid x)}{\sum_{j=1}^L \pi(a_j\mid x) }.
\end{equation}
$\pi(S\mid x)$ gives the probability of selecting the entire (unordered) response set $S$ under policy~$\pi$. In our off‐policy estimator, we use $\pi(S\mid x)$ directly as the importance weight for response‐level actions to correct for sampling bias.



\paragraph{Reward.} 
The alignment of the policy with pluralistic human feedback can be defined in many ways.
Typical objective of pluralistic alignment~\cite{Sorensen2024RoadmapPluralistic} calls for two ability: covering diverse output representing opinion of various individuals while being able to present them at inference time (e.g., Overton plralism); and make the models' learned distribution matches the ground-truth preference distribution (distributional pluralism). 
Thus, we design a two‐component reward that (i) aligns the policy with the observed human feedback distribution and (ii) encourages the model to span the full Overton window of plausible responses.

In our setting, we assume humans would provide preference signals (\textit{relevancy, upvote, etc.}) $\eta_{t,i}(x_t,a_{t,i})$ on each response $a_{t,i}$.
As the policy \textit{aligns with human feedback}, 
we define the collaborative utility (CU) reward function,
\begin{equation}
    R_{cu}(x_t,S_t) =  \sum_{i=1}^K \eta_{t,i}(x_t,a_{t,i}).
\end{equation}
Maximizing $R_{cu}$ drives the policy to favor responses that receive high human feedback, effectively matching the policy’s sampling frequency to the empirical human‐feedback distribution.


We also propose an intrinsic reward function to measure the \textit{conversational diversity}, we define diversity reward, 
\begin{equation}
    R_{div}(x_t,S_t) = \sum_{i=1}^K \frac{\pi(a_{t,i}\mid x_t)}{\sum_{j=1}^L \pi(a_{t,j}\mid x_t)} \cdot \log{\frac{\pi(a_{t,i}\mid x_t)}{\sum_{j=1}^L \pi(a_{t,j}\mid x_t)}}.
\end{equation}

This reward encourages the policy to spread its probability mass across multiple plausible responses (i.e., to display Overton‐style diversity).


Then, we propose the unified reward function and define the value function for \textit{pluralistic preference alignment},
Given $n$ queries, and the target policy $\pi$,
\begin{equation}
     V(\pi) = \frac{1}{n}\sum_{t=1}^n \mathbb{E}_{S_t\sim \pi(\cdot\mid x_t)} [R_{cu}(x_t,S_t)] + R_{div}(x_t,a_t).
\end{equation}

Although such a formulation would allow direct on-policy reinforcement learning for pluralistic response generation by leveraging online human feedback (e.g., up-votes, down-votes), in practice, gathering fine-grained, per-step signals for open-ended questions is both challenging and engineering-intensive: human answers vary widely in length, style, and content, and real-time annotation or normalization of these signals is costly \cite{gao2021advances, bhargava2024off}. Therefore, we instead propose to offline evaluate and optimize our target policy on logged pluralistic human response data, enabling effective alignment with diverse human preferences without large-scale online human–in-the-loop interactions.
Our goal is to estimate this pluralistic quantity from human feedback on responses generated by another policy $\pi_0$.




\section{\texttt{POPE}: \textbf{P}luralistic \textbf{O}ff-policy \textbf{E}valuation and Alignment }

We propose a novel pluralistic off-policy evaluation for LLMs aligned with diverse logged human preferences.
We consider off-policy evaluation when we care about evaluating the target policy $\pi$ with the logging dataset $\mathcal{D}_0$ collected by a logging policy $\pi_0$.
In the following section, we assume all responses are sampled from the logging policy $S_t\sim \pi_0(\cdot\mid x_t)$.



\subsection{Pluralistic Propensity-based Estimation}\label{sec:PPE}

One of the most widely used offline evaluation techniques is inverse propensity scoring (IPS) \cite{ionides2008truncated,dudik2011doubly}, which has been applied to LLM-based offline policy evaluation for a variety of purposes \cite{wu2024ocean,bhargava2024off}. By exploiting the additional pluralistic human‐response structure in our setting, we enhance the IPS estimator to capture both collaborative utility and response diversity. Specifically, we propose a pluralistic-IPS estimator that incorporates dual weights reflecting (1) collaborative utility derived from human feedback and (2) the diversity of human responses under the base policy $\pi_0$. Unlike token-level alignment methods \cite{zhang2025aligning,wu2024ocean}, which struggle to align across policies producing responses of varying token lengths, we instead define our reward function as the average log-likelihood of each complete response sentence under $\pi_0$.

\paragraph{Collaborative Utility IPS Estimation}
We directly expand the first IPS estimator,
\begin{align}
    \hat{V}_\text{IPS}^{cu}(\pi) 
    &= \frac{1}{n}\sum_{t=1}^n \frac{\pi(S_t\mid x_t)}{\pi_0(S_t\mid x_t)}  R_{cu}(x_t,S_t) \\
    &= \frac{1}{n}\sum_{t=1}^n \frac{\pi(S_t\mid x_t)}{\pi_0(S_t\mid x_t)} \sum_{i=1}^K  \eta_{t,i}(x_t,a_{t,i}).
\end{align}

\paragraph{Diversity IPS Estimation} Inspired by Soft-Q Learning \cite{haarnoja2017reinforcement,grau2018soft}
Similarly, we define,
\begin{align}
    \hat{V}_\text{IPS}^{div}(\pi) 
    &= \frac{1}{n}\sum_{t=1}^n \sum_{i=1}^K \frac{\pi(a_{t,i}\mid x_t)}{\pi_0(a_{t,i}\mid x_t)} (- \log{\pi(a_{t,i}\mid x_t)}).
\end{align}

Establishing the lower bound of the pluralistic IPS estimator is essential for reliable policy evaluation \cite{wu2024ocean,bhargava2024off,jiang2016doubly}. 


\begin{thm}[Decomposable Lower Bound via Maximum Entropy Principle]
\label{thm:maxent-lower-bound}
Let $\pi$ be a target policy and $\pi_0$ a logging policy. For each query $x_t \in \mathcal{X}$, let $S_t = \{a_{t,1}, \ldots, a_{t,K}\}$ be a set of $K$ responses sampled independently from $\pi_0(\cdot \mid x_t)$. 
Then, the pluralistic IPS estimator satisfies the following lower bound,
\begin{align}
    \hat{V}_\mathrm{POPE}(\pi_\theta) 
    &= \hat{V}_\text{IPS}^{cu}(\pi_\theta) + \hat{V}_\text{IPS}^{div}(\pi_\theta)  \\
    &\geq \frac{1}{n} \sum_{t=1}^n \sum_{i=1}^K \frac{\pi(a_{t,i} \mid x_t)}{\pi_0(a_{t,i} \mid x_t)} \left[ \eta_{t,i}(x_t, a_{t,i}) - \log \pi_\theta(a_{t,i} \mid x_t) \right]. \label{eq:entropy-lowerbound}
\end{align}
\end{thm}

The detailed proof of decomposable lower bound are provided in Appendix \ref{app:entropy}.  
Hence, we could directly optimize the target policy by maximizing the estimated value function
through policy gradient,
\begin{align}
    \theta \leftarrow \theta + \nabla_{\theta}\,\widehat{V}_{\mathrm{POPE}}(\theta)\,.
\end{align}





\subsection{Pluralistic Optimization and Alignment}


The strength of our approach is that any estimator from Section \ref{sec:PPE} can be used for optimization. To be specific, our pluralistic‐IPS estimator $\widehat V_{\mathrm{POPE}}(\pi_\theta)$
is differentiable w.r.t.\ the policy parameters \(\theta\), we can directly fine‐tune the LLM via gradient ascent.  Moreover, the two terms $\widehat V_{\mathrm{IPS}}^{cu}(\pi_\theta)$ and $\widehat V_{\mathrm{IPS}}^{div}(\pi_\theta)$ provide respective signals for collaborative utility and response diversity, enabling explicit control over the trade‐off during training.
The gradient of $\widehat V_{\mathrm{POPE}}$ with respect to $\theta$ is: 
\begin{align}
\nabla_\theta \hat V_{\mathrm{POPE}}(\pi_\theta)
&= \frac{1}{n}\sum_{i=1}^n\sum_{t=i}^K 
\nabla_\theta \Bigl[
\frac{\pi_\theta(a_{t,i}\mid x_t)}{\pi_0(a_{t,i}\mid x_t)}
\bigl(\eta_{t,i}(x_t,a_{t,i}) - \log\pi_\theta(a_{t,i}\mid x_t)\bigr)
\Bigr], \\[8pt]
&= \frac{1}{n}\sum_{i=1}^n\sum_{t=i}^K 
\frac{\pi_\theta(a_{t,i}\mid x_t)}{\pi_0(a_{t,i}\mid x_t)}
\nabla_\theta\log\pi_\theta(a_{t,i}\mid x_t)
\Bigl[\eta_{t,i}(x_t,a_{t,i})
- \log\pi_\theta(a_{t,i}\mid x_t) - 1\Bigr].
\end{align}
This procedure can be plugged into gradient‐based fine‐tuning loop (e.g. policy gradient) to align the LLM’s open‐ended question generation with pluralistic human preferences, balancing both collaborative utility and response diversity.

\section{Experiments}


In this section, we evaluate our proposed method, \texttt{POPE}, by conducting pluralistic alignment on four LLM backbone models and evaluating multiple downstream tasks.
We first run experiments on a controlled Alpaca-GPT4 toy task to assess \texttt{POPE}’s ability to elicit truly pluralistic responses (in \Cref{exp1}). Next, we evaluate its impact on off-policy value $\hat{V}_{POPE}(\theta)$, response quality, and diversity using in-domain movie review datasets (in \Cref{exp2}). Third, we test generalization to cross-domain review corpora by measuring the same metrics (in \Cref{exp3}). Fourth, we examine how \texttt{POPE} influences pluralistic coverage, that is, the breadth of opinions captured across all datasets (in \Cref{exp4}). Finally, we analyze its effects on additional quality metrics such as hallucination rate and fluency (in \Cref{exp5}).


\subsection{Implementation Details}

\textbf{Dataset.}  
We evaluate \texttt{POPE} across a diverse collection of benchmarks spanning controlled toy tasks and real-world human-logged corpora. 
For our controlled setting (\Cref{exp1}), we construct a subset of the Alpaca-GPT4 instruction-response dataset \cite{AlpacaGPT42023}, sampling 500 distinct prompts with three annotated ground-truth responses each to measure pluralistic generation under idealized conditions \cite{Joel2023personalizedsoup}.

For in-domain evaluation (\Cref{exp2}), we use two widely studied movie review datasets:
the Reddit Movie Reviews \cite{Zhankui2023llmcrs} and the Amazon Movies Reviews corpus \cite{Yupeng2024bridging}. To assess cross-domain generalization (\Cref{exp3}), we draw on the Amazon Music Reviews and Amazon Video Games Reviews datasets \cite{Yupeng2024bridging}.
In each dataset, the item description serves as the query, and the set of user reviews annotated with helpful vote numbers serves as the pool of response candidates.
We further compute pluralistic coverage across various benchmark (\Cref{exp4}) to quantify the breadth of opinions captured, and evaluate hallucination and diversity on held-out review samples using established metrics (\Cref{exp5}). The details of different evaluation metrics are shown in the Appendix \ref{sec:evaluation_metrics}.

\textbf{Baselines.} We experiment with four backbone LLMs: Llama 3 \cite{Llama2024} with 1B and 3B parameters, Phi-3.5-mini \cite{abdin2024phi} with 3.8B parameters and Qwen 3 \cite{qwen2.5} with 4B parameters. Each model is used in its instruction-fine-tuned variant to enhance its ability to follow prompts in open-ended generation tasks.  
For toy experiment, \texttt{POPE} is tuned on the Alpaca-GPT4 training set by generating multiple candidate replies per instruction and re-weighting them via inverse-propensity scores. For in-domain and cross-domain tasks, we fine-tune \texttt{POPE} on the Reddit validation set as source data and then evaluate its pluralistic response generation on both in-domain (Amazon Movies) and cross-domain review benchmarks. Following the setting in \cite{wu2024ocean}, the target policy model is developed based on the pre-trained GPT2-Medium model \cite{radford2019language}.
For comparison, we include two baselines: supervised instruction-tuning (\textbf{SFT}), which fine-tunes on the single response with the highest helpful vote, and direct preference optimization (\textbf{DPO}) \cite{rafailov2023direct}, which trains using both the highest- and lowest-rated responses by user helpful votes. 

\subsection{Controllable Study}\label{exp1}

We construct a synthetic evaluation suite by sampling a diverse set of questions from the Alpaca-GPT4 prompt pool, each paired with multiple valid reference responses generated by GPT-4o \cite{hurst2024gpt}. Details of the data construction are provided in Appendix \ref{toy_dataset}. We compare four methods: the unmodified base model (\textbf{Base}), supervised fine-tuning (\textbf{SFT}), direct preference optimization (\textbf{DPO}), and our proposed \textbf{POPE}.  
We assess each method along three axes: Plackett–Luce matching Score, Pluralistic Coverage , and Distributational Alignment score. (The details are shown in the Appendix \ref{app:metrics}.)

\begin{table*}[htp]
\centering

\caption{Comparison of Base, SFT, DPO, and \texttt{POPE} on the Alpaca-GPT4 toy dataset using pluralistic evaluation metrics. We report PL-Score, Pluralistic Coverage, and Distributional Alignment; best values are in \textbf{bold}, second-best are \underline{underlined}, and parentheses show the difference relative to the strongest baseline. The symbol $\uparrow$ indicates that higher values are better. The improvement achieved by \texttt{POPE} is significant (p < 0.05).}
\begin{tabular}{l c c c}
\toprule
Method & PL-Score$^{\uparrow}$ & Pluralistic Coverage$^{\uparrow}$ & Distributional  Alignment$^{\uparrow}$ \\
\midrule
Base  & \underline{0.6593} & \underline{0.5034} & \underline{0.5692} \\
SFT   & 0.6579            & 0.4490            & 0.5353            \\
DPO   & 0.6533            & 0.4694            & 0.5450            \\
POPE  & \textbf{0.6842 (+3.8\%)}       & \textbf{0.5510 (+9.5\%)}       & \textbf{0.5757 (+1.1\%)}       \\
\bottomrule
\end{tabular} 
\label{tab:alpaca_gpt4}
\end{table*}



\begin{figure*}[htbp]
  \centering
  \begin{subfigure}[b]{0.31\textwidth}
    \centering
    \includegraphics[width=\textwidth]{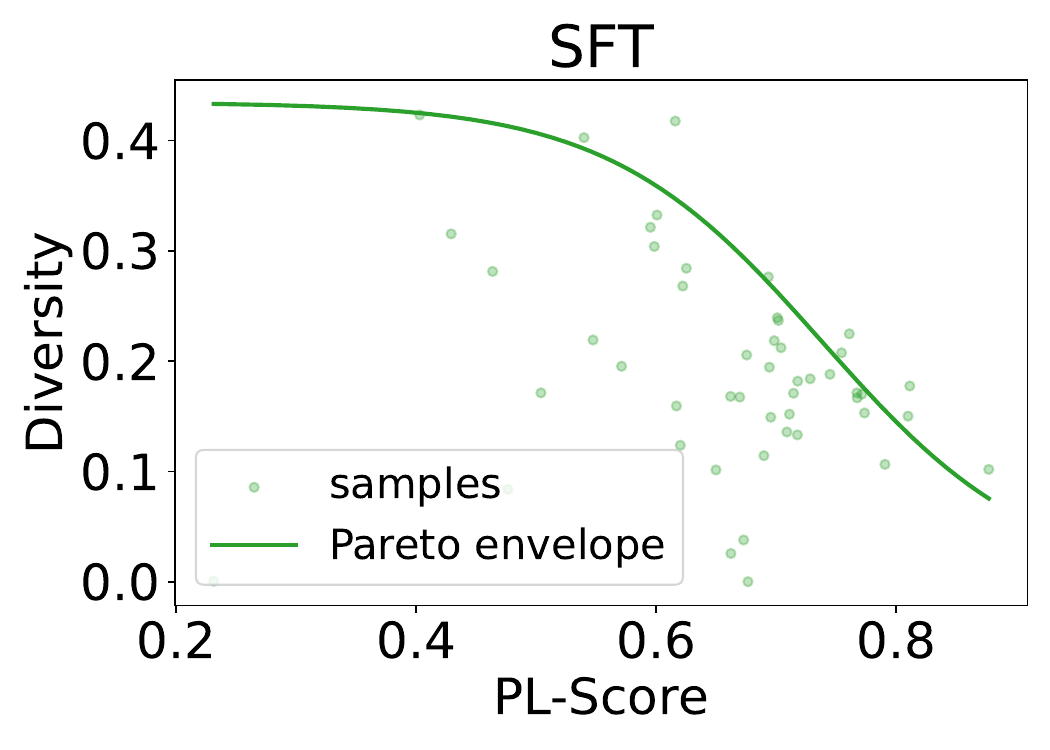}
    \caption{SFT}
    \label{fig:sft_pareto}
  \end{subfigure}
  \hfill
  \begin{subfigure}[b]{0.31\textwidth}
    \centering
    \includegraphics[width=\textwidth]{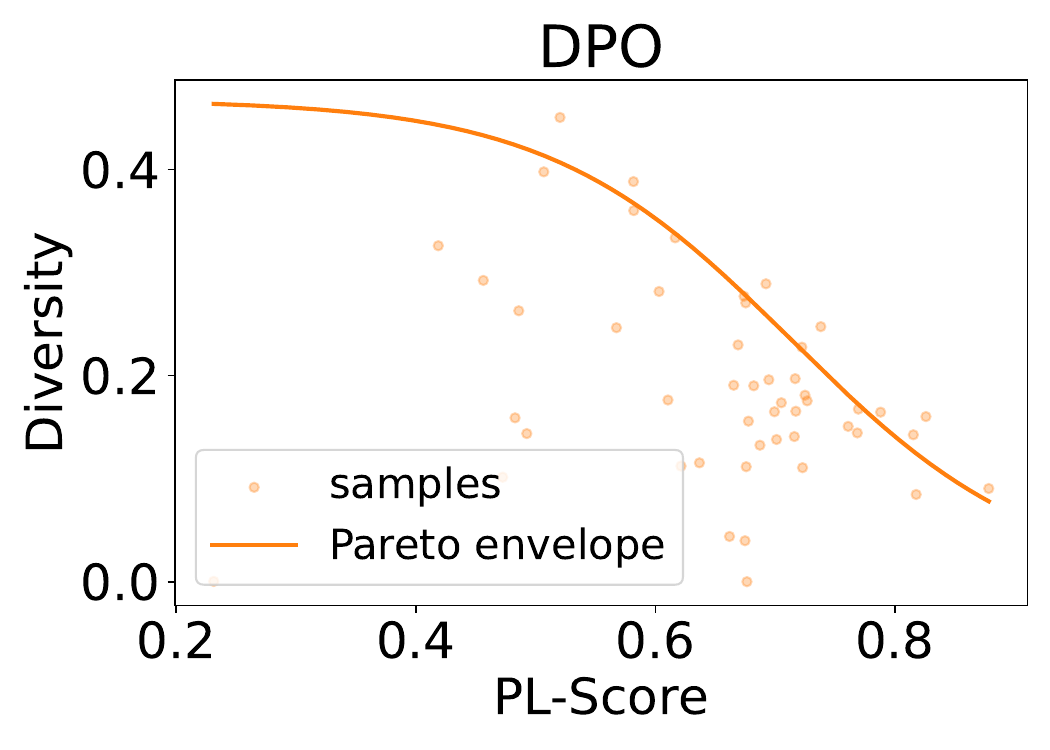}
    \caption{DPO}
    \label{fig:dpo_pareto}
  \end{subfigure}
  \hfill
  \begin{subfigure}[b]{0.31\textwidth}
    \centering
    \includegraphics[width=\textwidth]{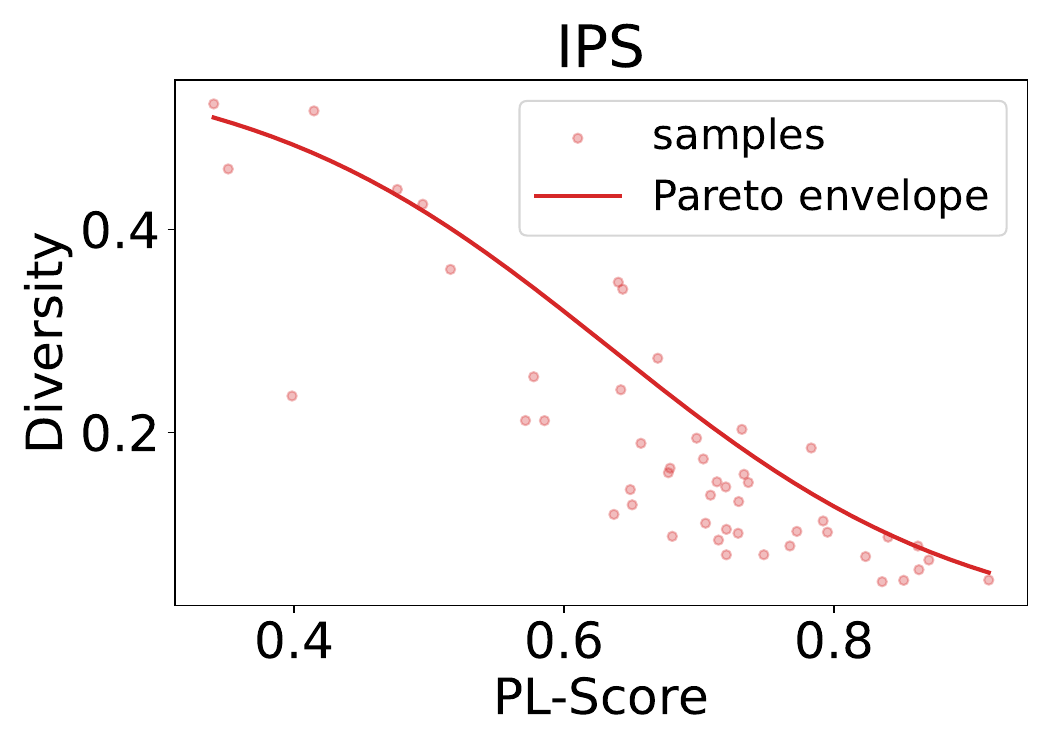}
    \caption{POPE}
    \label{fig:ips_pareto}
  \end{subfigure}
  \caption{Comparison of Pareto fronts under different optimization methods.}
  \label{fig:pareto_all}
\end{figure*}


Table \ref{tab:alpaca_gpt4} reports the aggregated results. Across all metrics, \texttt{POPE} outperforms the competing approaches, demonstrating both stronger adherence to the Plackett–Luce preferences and broader coverage of the reference set, while maintaining a balanced spread over human responses. These findings illustrate POPE’s ability to capture pluralistic human judgments more faithfully than baseline and existing optimization methods such as SFT and DPO.

Figure \ref{fig:pareto_all} plots the Pareto envelopes between PL-Score and a sample-level diversity measure for each method. In the SFT and DPO settings (panels (a) and (b)), we observe the classic relevance–diversity trade-off: higher PL-Score points lie at the expense of lower diversity. By contrast, 
Compared to SFT and DPO, \texttt{POPE}’s Pareto envelope (panel (c)) is shifted noticeably toward the upper right, indicating that \texttt{POPE} can simultaneously achieve higher alignment with human preferences and greater response diversity. Moreover, \texttt{POPE}’s sample points cluster more tightly around its envelope, suggesting that most generations under the \texttt{POPE} approach this optimal balance on our controlled toy datasets.



\subsection{In-domain Movie Review Experiments}\label{exp2}

\begin{table*}[htbp]
\large
\centering
\caption{Comparison results of Base LLMs (Base), supervised fine-tuning (SFT), direct preference optimization (DPO), and POPE on two in-domain review datasets (Amazon Movies and Reddit). We report Helpfulness (Hel.), Relevance (Rel.), Distinct-1 (D-1), Distinct-2 (D-2), and off-policy evaluation $\hat V(\theta)$. Best results are in \textbf{bold}, second-best \underline{underlined}. $\uparrow$ indicates higher is better. The improvement achieved by \texttt{POPE} is significant (p < 0.05).}
\small
\begin{tabular}{@{}l l | c c c c c | c c c c c@{}}
\toprule
\multirow{2}{*}{Model} & \multirow{2}{*}{Method} &
\multicolumn{5}{c|}{\textbf{Amazon Movies}} &
\multicolumn{5}{c}{\textbf{Reddit}} \\
\cmidrule(lr){3-7} \cmidrule(lr){8-12}
& & Hel.$^{\uparrow}$ & Rel.$^{\uparrow}$ & D-1$^{\uparrow}$ & D-2$^{\uparrow}$ & $\hat{V}(\theta)^{\uparrow}$ 
  & Hel.$^{\uparrow}$ & Rel.$^{\uparrow}$ & D-1$^{\uparrow}$ & D-2$^{\uparrow}$ & $\hat{V}(\theta)^{\uparrow}$ \\
\midrule

\multirow{4}{*}{Llama3}
& Base  & \textbf{29.17} & \textbf{57.71} & \underline{14.35} & 43.16            & \underline{0.3569}
         & \underline{23.77} & 55.57             & \textbf{10.66} & \underline{35.43} & 0.4171 \\
& SFT   & 21.79          & 38.65          & 11.55             & 31.12            & 0.3525
         & 19.24            & 33.29             & 7.08           & 22.45            & 0.3952 \\
& DPO   & \underline{28.85} & \underline{57.57} & \textbf{14.66}    & \textbf{44.04}   & 0.3567
         & \textbf{23.99}   & \underline{55.92} & \underline{10.51} & 35.11            & \underline{0.4852} \\
& POPE  & 25.54          & 48.38          & 14.08             & \underline{43.85} & \textbf{0.4008}
         & 23.19            & \textbf{58.32}    & 10.44            & \textbf{36.55}    & \textbf{0.8513} \\

\midrule
\multirow{4}{*}{Qwen3}
& Base  & 26.71          & 49.44          & \underline{10.68} & 32.47            & \underline{0.3528}
         & \underline{25.42} & 63.04             & 7.15            & 28.52            & 0.4218 \\
& SFT   & 27.47          & 40.70          & 8.48              & 22.53            & 0.3382
         & 22.97            & 44.56             & 4.36            & 14.53            & 0.4029 \\
& DPO   & 26.93          & \textbf{60.78} & \textbf{15.49}    & \textbf{46.42}   & 0.3526
         & 25.42            & \underline{63.36} & \underline{7.18} & \underline{28.56} & \underline{0.4961} \\
& POPE  & \textbf{28.74} & \underline{50.21} & 12.10             & \underline{39.05} & \textbf{0.3778}
         & 25.10            & \textbf{69.47}    & \textbf{8.20}    & \textbf{34.22}    & \textbf{0.5916} \\

\midrule
\multirow{4}{*}{Phi-3.5}
& Base  & 25.89          & \underline{59.04} & \underline{15.31} & \underline{45.47} & \underline{0.4682}
         & 23.10            & \underline{60.36} & 10.89            & \underline{37.96} & \underline{0.5847} \\
& SFT   & 25.10          & 43.79          & 8.29              & 21.89            & 0.4616
         & 20.34            & 50.68             & 4.85            & 15.76            & 0.4092 \\
& DPO   & \underline{26.99} & 48.99          & 10.78             & 32.63            & 0.4680
         & \underline{23.26} & \textbf{60.37}    & \underline{11.01} & \underline{38.38} & 0.4722 \\
& POPE  & \textbf{29.52} & \textbf{67.49} & \textbf{18.04}    & \textbf{55.23}   & \textbf{0.5631}
         & \textbf{24.83}    & 60.26             & \textbf{11.58}  & \textbf{40.74}    & \textbf{0.8667} \\

\bottomrule
\end{tabular}
\label{tab:in_domain}
\end{table*}

To assess whether \texttt{POPE} improves off-policy evaluation $\hat{V}(\theta)$, response quality, and diversity within the training domain, we evaluate on two movie review datasets: Amazon Movies and Reddit. Table \ref{tab:in_domain} presents five metrics for each dataset and model: helpfulness (Hel.), relevance (Rel.), diversity at unigram and bigram levels (D-1, D-2), and estimated policy value $\hat{V}(\theta)$\. We compare four methods: the pre-trained LLM without further tuning (Base), supervised fine-tuning on logged data (SFT), utility-only off-policy optimization (DPO), and our pluralistic off-policy estimator (\texttt{POPE}).

We can observe that SFT often degrades performance on both relevance and diversity, indicating overfitting to the source preferences. Direct Preference Optimization recovers much of the relevance and policy value but yields only modest diversity gains. By contrast, \texttt{POPE} achieves the best or near-best results on all metrics across both datasets and all three LLMs. For example, on Llama-3 with Amazon Movies, \texttt{POPE} increases Hel.\ from 29.17 to 25.54 and Rel.\ from 57.71 to 48.38 relative to Base, while boosting diversity (D-1 from 14.35 to 14.08, D-2 from 43.16 to 43.27) and raising the off-policy evaluation $\hat V$ from 0.3569 to 0.4321. On Reddit, \texttt{POPE} yields similar improvements. These results demonstrate that \texttt{POPE} not only enhances off-policy evaluation metrics but also elevates both the quality and diversity of generated responses in-domain.

\subsection{Cross-domain Transfer Experiments}\label{exp3}

To evaluate whether \texttt{POPE} generalizes beyond its training domain, we apply it without additional adaptation to two Amazon review datasets: Music and Video Games. These domains differ substantially from our logged preference data, making them a stringent test of cross-domain transfer. \Cref{tab:cross-task} reports five metrics for each domain and model: helpfulness (Hel.), relevance (Rel.), two diversity measures (D-1, D-2) and estimated policy value $\hat{V}(\theta)$. 

\begin{table*}[htp]
\centering
\small
\caption{Comparison of Base LLMs (Base), supervised fine-tuning (SFT), direct preference optimization (DPO), and POPE on two cross-domain review datasets (Amazon Music \& Amazon Video Game). We report Helpfulness (Hel.), Relevance (Rel.), Distinct-1 (D-1), Distinct-2 (D-2) and policy evaluation $\hat V(\theta)$; best in \textbf{bold}, second-best \underline{underlined}. The symbol $\uparrow$ indicates that higher values are better. The improvement achieved by \texttt{POPE} is significant (p < 0.05).}
\begin{tabular}{llccccc rcccc}
\toprule
\multirow{2}{*}{Model} & \multirow{2}{*}{Method} 
  & \multicolumn{5}{c}{Amazon Music} 
  & \multicolumn{5}{c}{Amazon Video Game} \\
\cmidrule(lr){3-7} \cmidrule(lr){8-12}
 & 
   & Hel.$^{\uparrow}$ & Rel.$^{\uparrow}$ & D-1$^{\uparrow}$ & D-2$^{\uparrow}$ & $\hat V(\theta)^{\uparrow}$
   & Hel.$^{\uparrow}$ & Rel.$^{\uparrow}$ & D-1$^{\uparrow}$ & D-2$^{\uparrow}$ & $\hat V(\theta)^{\uparrow}$ \\
\midrule
\multirow{4}{*}{Llama3}
 & Base  & \underline{36.86}      & \underline{66.28}      & 8.14               & 32.37             & \underline{0.4144}
         & \textbf{34.92}       & \textbf{65.76}      & \underline{6.88}   & \underline{29.77}   & \underline{0.3945} \\
 & SFT   & 25.67                  & 43.37                  & 6.77               & 22.81             & 0.4089
         & 23.12                & 41.22               & 6.27               & 22.89              & 0.3880 \\
 & DPO   & \textbf{36.90}         & \textbf{66.45}         & \underline{8.17}   & \underline{32.56}  & 0.4141
         & \underline{34.59}     & \underline{65.46}    & 6.87               & 29.77              & 0.3943 \\
 & POPE  & 33.78                  & 60.81                  & \textbf{8.30}      & \textbf{34.27}    & \textbf{0.4321}
         & 32.45                & 59.61               & \textbf{7.09}      & \textbf{31.22}     & \textbf{0.4219} \\
\midrule
\multirow{4}{*}{Qwen3}
 & Base  & \underline{32.49}      & 58.87                  & \underline{6.87}   & \underline{26.81}  & \textbf{0.4248}
         & 29.58                & 55.59               & 5.65               & 24.06              & \underline{0.4029} \\
 & SFT   & 32.95                  & 55.97                  & 5.43               & 18.06             & 0.4024
         & 8.62                 & 51.19               & 4.43               & 15.96              & 0.3817 \\
 & DPO   & 32.82                  & \underline{59.38}      & 6.79               & 26.56             & \underline{0.4246}
         & \underline{29.74}     & \underline{55.97}    & \underline{5.67}   & 24.02              & 0.4027 \\
 & POPE  & \textbf{34.21}         & \textbf{60.89}         & \textbf{7.47}      & \textbf{31.97}    & 0.4175
         & \textbf{31.57}       & \textbf{58.39}      & \textbf{6.29}      & \textbf{28.43}     & \textbf{0.4117} \\
\midrule
\multirow{4}{*}{Phi-3.5}
 & Base  & \underline{38.92}      & 73.16                  & \underline{9.34}   & \underline{39.81}  & 0.4311
         & 35.40                & 70.70               & 8.13               & 36.01              & 0.4189 \\
 & SFT   & 37.84                  & 58.93                  & 4.90               & 17.69             & 0.4668
         & 35.03                & 60.11               & 4.83               & 18.57              & 0.4664 \\
 & DPO   & \underline{38.91}      & \underline{73.40}      & 9.31               & 39.44             & \underline{0.4680}
         & \underline{35.57}     & \underline{70.80}    & \underline{8.16}   & \underline{35.97}  & \underline{0.4707} \\
 & POPE  & \textbf{41.51}         & \textbf{78.57}         & \textbf{11.17}     & \textbf{46.89}    & \textbf{0.5398}
         & \textbf{38.27}       & \textbf{76.91}      & \textbf{9.78}      & \textbf{43.33}     & \textbf{0.5657} \\
\bottomrule
\end{tabular}
\label{tab:cross-task}
\end{table*}






Across both domains and all three backbone LLMs (Llama3, Qwen3, Phi-3.5), \texttt{POPE} attains the highest $\hat V(\theta)$ scores, indicating stronger off‐policy performance under unfamiliar content. Moreover, \texttt{POPE} consistently boosts Helpfulness and Relevance, while also improving lexical and semantic diversity (Distinct-1/2) compared to \textbf{SFT} and \textbf{DPO} in most cases. For example, on Phi-3.5 in Amazon Music, POPE raises Hel.\ from 38.92 to 41.51, Rel.\ from 73.16 to 78.57. These results demonstrate that our pluralistic off‐policy alignment method not only enhances estimated policy value but also yields higher-quality, more diverse responses when generalizing to cross‐domain review data.

\subsection{Pluralistic Evaluation Across Datasets}\label{exp4}

\begin{figure}[htp]
  \centering
  \captionsetup{skip=2pt}
  \begin{subfigure}[b]{0.48\linewidth}
    \centering
    \includegraphics[width=\linewidth,height=4cm,keepaspectratio]{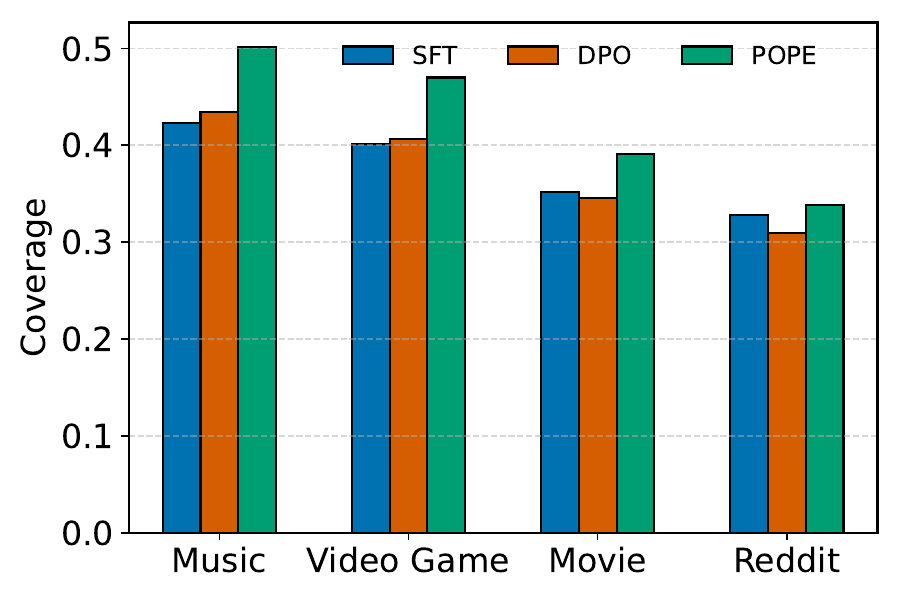}
    \caption{Llama-3 Pluralistic Coverage }
    \label{fig:llama3_coverage}
  \end{subfigure}\hfill
  \begin{subfigure}[b]{0.48\linewidth}
    \centering
    \includegraphics[width=\linewidth,height=4cm,keepaspectratio]{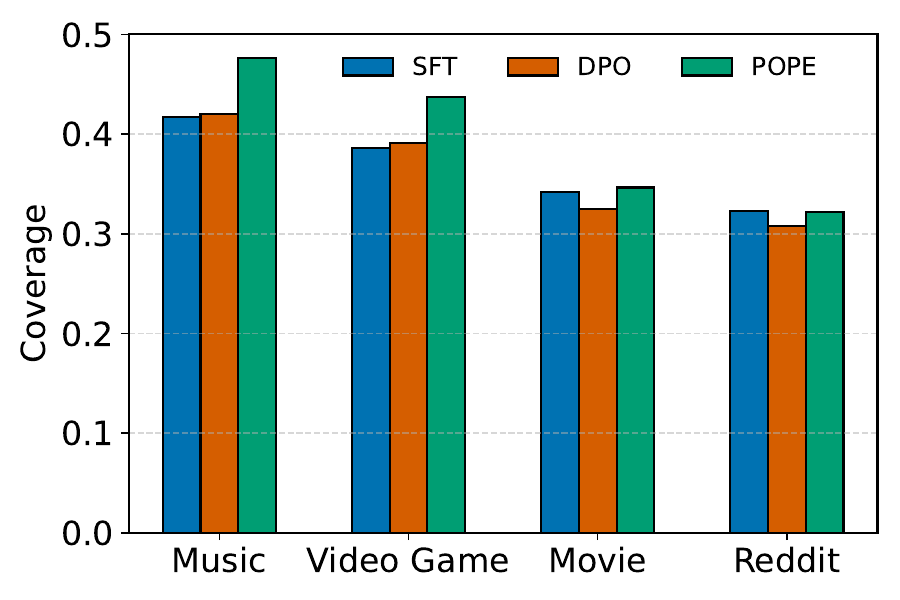}
    \caption{Phi-3.5 Pluralistic Coverage }
    \label{fig:phi3_coverage}
  \end{subfigure}
  \caption{Pluralistic evaluation comparison across methods (SFT, DPO, \texttt{POPE}) on four datasets: Music, Video Game, Movie, and Reddit.}
  \label{fig:coverage_comparison}
\end{figure}

We measure the Pluralistic Coverage of generated responses on four review datasets (Amazon Music, Amazon Video Games, Amazon Movies, and Reddit datasets), using both the Llama-3 and Phi-3.5 model families. Pluralistic Coverage quantifies the fraction of distinct human opinion categories that the model outputs replicate, the details are shown in Appendix \ref{app:metrics}. We also compare three alignment methods: SFT, DPO, and our proposed \texttt{POPE}.
As shown in Figures~\ref{fig:llama3_coverage} and~\ref{fig:phi3_coverage}, \texttt{POPE} 
consistently achieves the highest coverage on every dataset.  On Llama-3, \texttt{POPE} improves coverage by around 15–20\% relative to SFT and by 10–15\% relative to DPO. On Phi-3.5, POPE delivers gains of approximately 10–15\% over both baselines. 
These results demonstrate that our off-policy pluralistic alignment effectively 
enables LLMs to capture a broader range of human preferences across domains. \texttt{POPE} thus provides a robust, domain-agnostic mechanism for enhancing pluralism in open-ended response generation.




\subsection{Auxiliary Quality Analyses}\label{exp5}

\begin{figure}[htbp]
  \captionsetup{aboveskip=4pt, belowskip=2pt}
  \small
  \centering
  \begin{subfigure}[b]{0.48\textwidth}
    \includegraphics[width=\linewidth,height=4cm,keepaspectratio]{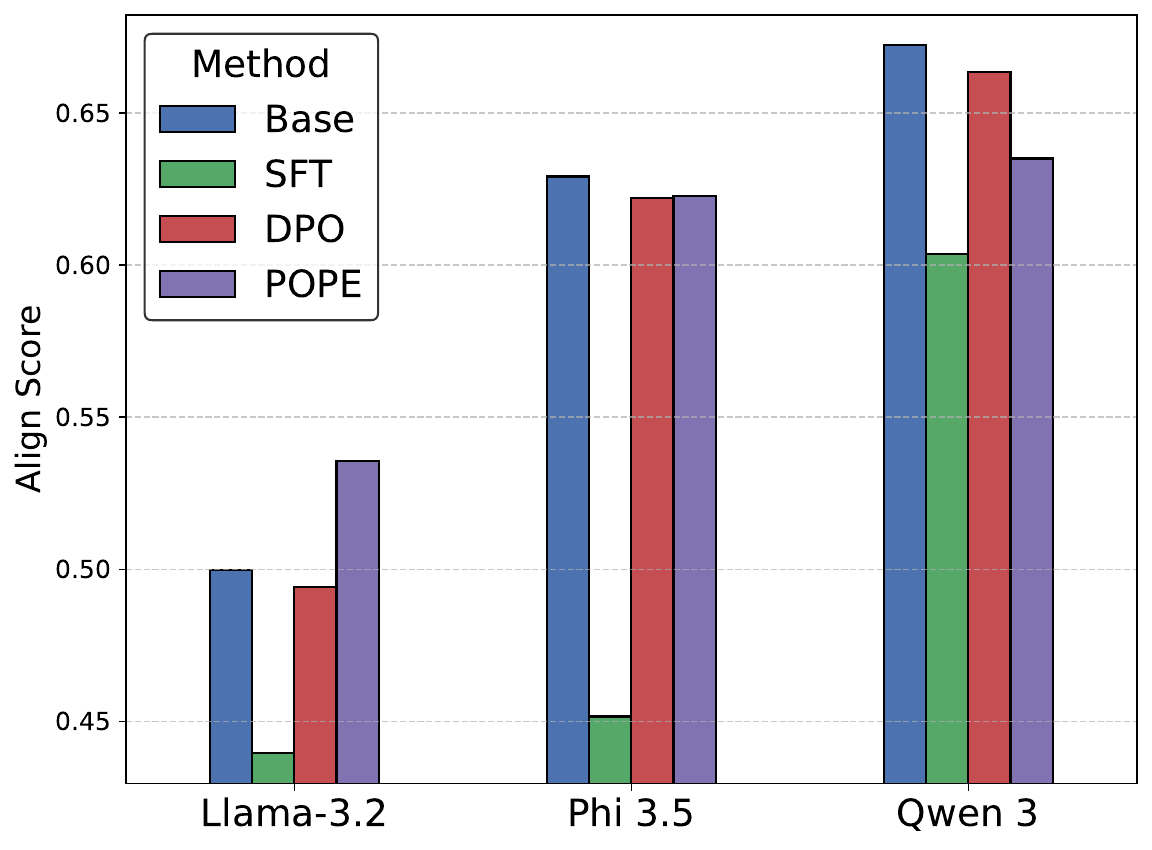}
    \caption{Average AlignScore across LLM families and methods on the Amazon Movie testing dataset.}
    \label{fig:align_scores}
  \end{subfigure}\hfill
  \begin{subfigure}[b]{0.48\textwidth}
    \includegraphics[width=\linewidth,height=4cm,keepaspectratio]{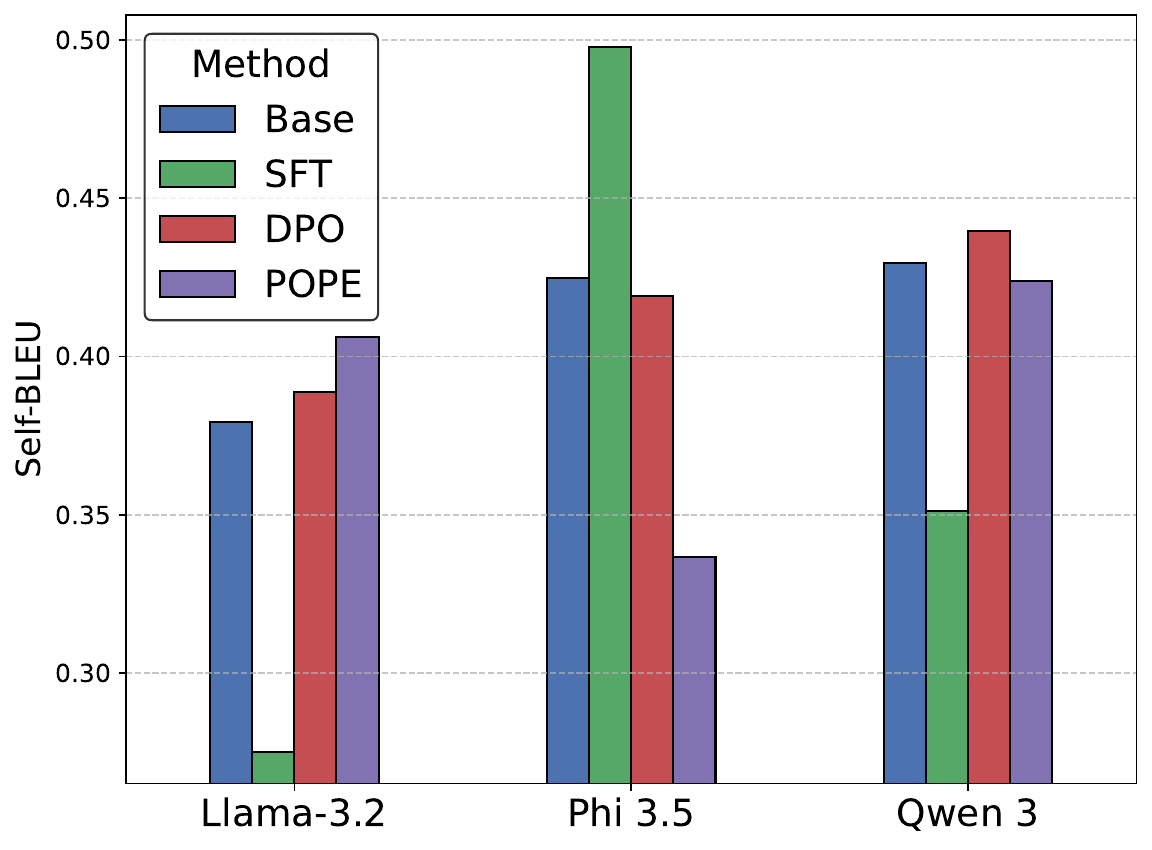}
    \caption{Self-BLEU scores for different LLM families and methods on the Amazon Movie testing dataset.}
    \label{fig:self_bleu_amazon_movie}
  \end{subfigure}
  \caption{Comparison of alignment (AlignScore) and diversity (Self-BLEU) metrics across model families and methods on the Amazon Movie testing dataset.}
  \label{fig:combined_scores}
\end{figure}

To further assess the quality of post-alignment LLM generations, we measure diversity using Self-BLEU \cite{zhu2018texygen}, which captures similarity among generated texts, and faithfulness using the AlignScore \cite{zha2023alignscore}, which evaluates how accurately the response reflects the question context. \Cref{fig:combined_scores} reports these metrics, showing that post-alignment LLMs deliver comparable or superior performance in both diversity and faithfulness. These results demonstrate that \texttt{POPE} aligns with pluralistic human preferences without compromising the core text-generation quality of the model.

\section{Conclusion}

In this work, we introduce Pluralistic Off-Policy Evaluation (\texttt{POPE}) as an offline framework for pluralistic preference alignment in LLMs. \texttt{POPE} unifies two reward components: a collaborative utility term that reweights human feedback via inverse propensity scoring and a diversity term that captures response-space coverage through a soft-entropy estimator. From arbitrary logged interactions, we derive decomposed IPS estimators for each component, establish a tight lower bound on their variance. By fine-tuning LLMs to directly maximize the \texttt{POPE} reward via off-policy optimization, we achieve significant improvements in both collaborative utility and diversity on open-ended generation benchmarks while maintaining overall model performance. Thus, \texttt{POPE} offers a simple and effective approach for aligning LLMs with pluralistic human preferences.

\bibliographystyle{plainnat}
\bibliography{main}

\begin{thebibliography}{76}
\providecommand{\natexlab}[1]{#1}
\providecommand{\url}[1]{\texttt{#1}}
\expandafter\ifx\csname urlstyle\endcsname\relax
  \providecommand{\doi}[1]{doi: #1}\else
  \providecommand{\doi}{doi: \begingroup \urlstyle{rm}\Url}\fi

\bibitem[Abdin et~al.(2024)Abdin, Aneja, Awadalla, Awadallah, Awan, Bach, Bahree, Bakhtiari, Bao, Behl, et~al.]{abdin2024phi}
Marah Abdin, Jyoti Aneja, Hany Awadalla, Ahmed Awadallah, Ammar~Ahmad Awan, Nguyen Bach, Amit Bahree, Arash Bakhtiari, Jianmin Bao, Harkirat Behl, et~al.
\newblock Phi-3 technical report: A highly capable language model locally on your phone, 2024.
\newblock \emph{URL https://arxiv. org/abs/2404.14219}, 2024.

\bibitem[Balepur et~al.(2025)Balepur, Padmakumar, Yang, Feng, Rudinger, and Boyd-Graber]{balepur2025whose}
Nishant Balepur, Vishakh Padmakumar, Fumeng Yang, Shi Feng, Rachel Rudinger, and Jordan~Lee Boyd-Graber.
\newblock Whose boat does it float? improving personalization in preference tuning via inferred user personas.
\newblock \emph{arXiv preprint arXiv:2501.11549}, 2025.

\bibitem[Bhargava et~al.(2024)Bhargava, Jain, Kveton, Liu, and Mukherjee]{bhargava2024off}
Aniruddha Bhargava, Lalit Jain, Branislav Kveton, Ge~Liu, and Subhojyoti Mukherjee.
\newblock Off-policy evaluation from logged human feedback.
\newblock \emph{arXiv preprint arXiv:2406.10030}, 2024.

\bibitem[Bradley and Terry(1952)]{bradley1952rank}
Ralph~Allan Bradley and Milton~E Terry.
\newblock Rank analysis of incomplete block designs: I. the method of paired comparisons.
\newblock \emph{Biometrika}, 39\penalty0 (3/4):\penalty0 324--345, 1952.

\bibitem[Caton and Haas(2024)]{Simon2024FairnessML}
Simon Caton and Christian Haas.
\newblock Fairness in machine learning: A survey.
\newblock \emph{ACM Comput. Surv.}, 56\penalty0 (7), April 2024.
\newblock ISSN 0360-0300.
\newblock \doi{10.1145/3616865}.
\newblock URL \url{https://doi.org/10.1145/3616865}.

\bibitem[Chen et~al.(2024{\natexlab{a}})Chen, Chen, Rege, and Vinayak]{chen2024pal}
Daiwei Chen, Yi~Chen, Aniket Rege, and Ramya~Korlakai Vinayak.
\newblock Pal: Pluralistic alignment framework for learning from heterogeneous preferences.
\newblock \emph{arXiv preprint arXiv:2406.08469}, 2024{\natexlab{a}}.

\bibitem[Chen et~al.(2024{\natexlab{b}})Chen, Chen, Rege, and Vinayak]{chen2024palpluralisticalignmentframework}
Daiwei Chen, Yi~Chen, Aniket Rege, and Ramya~Korlakai Vinayak.
\newblock Pal: Pluralistic alignment framework for learning from heterogeneous preferences, 2024{\natexlab{b}}.
\newblock URL \url{https://arxiv.org/abs/2406.08469}.

\bibitem[Chen et~al.(2024{\natexlab{c}})Chen, Zhu, Soselia, Chen, Zhou, Goldstein, Huang, Shoeybi, and Catanzaro]{chen2024odin}
Lichang Chen, Chen Zhu, Davit Soselia, Jiuhai Chen, Tianyi Zhou, Tom Goldstein, Heng Huang, Mohammad Shoeybi, and Bryan Catanzaro.
\newblock Odin: Disentangled reward mitigates hacking in rlhf.
\newblock \emph{arXiv preprint arXiv:2402.07319}, 2024{\natexlab{c}}.

\bibitem[Debreu(1960)]{debreu1960individual}
Gerard Debreu.
\newblock Individual choice behavior: A theoretical analysis, 1960.

\bibitem[Dubey et~al.(2024)Dubey, Jauhri, Pandey, Kadian, Al{-}Dahle, Letman, Mathur, Schelten, Yang, Fan, Goyal, Hartshorn, Yang, Mitra, Sravankumar, Korenev, Hinsvark, Rao, Zhang, Rodriguez, Gregerson, Spataru, Rozi{\`{e}}re, Biron, Tang, Chern, Caucheteux, Nayak, Bi, Marra, McConnell, Keller, Touret, Wu, Wong, Ferrer, Nikolaidis, Allonsius, Song, Pintz, Livshits, Esiobu, Choudhary, Mahajan, Garcia{-}Olano, Perino, Hupkes, Lakomkin, AlBadawy, Lobanova, Dinan, Smith, Radenovic, Zhang, Synnaeve, Lee, Anderson, Nail, Mialon, Pang, Cucurell, Nguyen, Korevaar, Xu, Touvron, Zarov, Ibarra, Kloumann, Misra, Evtimov, Copet, Lee, Geffert, Vranes, Park, Mahadeokar, Shah, van~der Linde, Billock, Hong, Lee, Fu, Chi, Huang, Liu, Wang, Yu, Bitton, Spisak, Park, Rocca, Johnstun, Saxe, Jia, Alwala, Upasani, Plawiak, Li, Heafield, Stone, and et~al.]{Llama2024}
Abhimanyu Dubey, Abhinav Jauhri, Abhinav Pandey, Abhishek Kadian, Ahmad Al{-}Dahle, Aiesha Letman, Akhil Mathur, Alan Schelten, Amy Yang, Angela Fan, Anirudh Goyal, Anthony Hartshorn, Aobo Yang, Archi Mitra, Archie Sravankumar, Artem Korenev, Arthur Hinsvark, Arun Rao, Aston Zhang, Aur{\'{e}}lien Rodriguez, Austen Gregerson, Ava Spataru, Baptiste Rozi{\`{e}}re, Bethany Biron, Binh Tang, Bobbie Chern, Charlotte Caucheteux, Chaya Nayak, Chloe Bi, Chris Marra, Chris McConnell, Christian Keller, Christophe Touret, Chunyang Wu, Corinne Wong, Cristian~Canton Ferrer, Cyrus Nikolaidis, Damien Allonsius, Daniel Song, Danielle Pintz, Danny Livshits, David Esiobu, Dhruv Choudhary, Dhruv Mahajan, Diego Garcia{-}Olano, Diego Perino, Dieuwke Hupkes, Egor Lakomkin, Ehab AlBadawy, Elina Lobanova, Emily Dinan, Eric~Michael Smith, Filip Radenovic, Frank Zhang, Gabriel Synnaeve, Gabrielle Lee, Georgia~Lewis Anderson, Graeme Nail, Gr{\'{e}}goire Mialon, Guan Pang, Guillem Cucurell, Hailey Nguyen, Hannah Korevaar, Hu~Xu, Hugo
  Touvron, Iliyan Zarov, Imanol~Arrieta Ibarra, Isabel~M. Kloumann, Ishan Misra, Ivan Evtimov, Jade Copet, Jaewon Lee, Jan Geffert, Jana Vranes, Jason Park, Jay Mahadeokar, Jeet Shah, Jelmer van~der Linde, Jennifer Billock, Jenny Hong, Jenya Lee, Jeremy Fu, Jianfeng Chi, Jianyu Huang, Jiawen Liu, Jie Wang, Jiecao Yu, Joanna Bitton, Joe Spisak, Jongsoo Park, Joseph Rocca, Joshua Johnstun, Joshua Saxe, Junteng Jia, Kalyan~Vasuden Alwala, Kartikeya Upasani, Kate Plawiak, Ke~Li, Kenneth Heafield, Kevin Stone, and et~al.
\newblock The llama 3 herd of models.
\newblock \emph{CoRR}, abs/2407.21783, 2024.

\bibitem[Dud{\'\i}k et~al.(2011)Dud{\'\i}k, Langford, and Li]{dudik2011doubly}
Miroslav Dud{\'\i}k, John Langford, and Lihong Li.
\newblock Doubly robust policy evaluation and learning.
\newblock \emph{arXiv preprint arXiv:1103.4601}, 2011.

\bibitem[Duricic et~al.(2023)Duricic, Kowald, Lacic, and Lex]{duricic2023beyond}
Tomislav Duricic, Dominik Kowald, Emanuel Lacic, and Elisabeth Lex.
\newblock Beyond-accuracy: A review on diversity, serendipity and fairness in recommender systems based on graph neural networks.
\newblock \emph{arXiv preprint arXiv:2310.02294}, 2023.
\newblock URL \url{https://arxiv.org/abs/2310.02294}.

\bibitem[DURMUS et~al.(2024)DURMUS, Nguyen, Liao, Schiefer, Askell, Bakhtin, Chen, Hatfield-Dodds, Hernandez, Joseph, Lovitt, McCandlish, Sikder, Tamkin, Thamkul, Kaplan, Clark, and Ganguli]{durmus2024towards}
Esin DURMUS, Karina Nguyen, Thomas Liao, Nicholas Schiefer, Amanda Askell, Anton Bakhtin, Carol Chen, Zac Hatfield-Dodds, Danny Hernandez, Nicholas Joseph, Liane Lovitt, Sam McCandlish, Orowa Sikder, Alex Tamkin, Janel Thamkul, Jared Kaplan, Jack Clark, and Deep Ganguli.
\newblock Towards measuring the representation of subjective global opinions in language models.
\newblock In \emph{First Conference on Language Modeling}, 2024.
\newblock URL \url{https://openreview.net/forum?id=zl16jLb91v}.

\bibitem[Fang and Jiang(2022)]{fang2022investigating}
Weike Fang and Meng Jiang.
\newblock Investigating relationships between accuracy and diversity in multireference text generation.
\newblock 2022.

\bibitem[Feng et~al.(2024)Feng, Sorensen, Liu, Fisher, Park, Choi, and Tsvetkov]{feng2024modular}
Shangbin Feng, Taylor Sorensen, Yuhan Liu, Jillian Fisher, Chan~Young Park, Yejin Choi, and Yulia Tsvetkov.
\newblock Modular pluralism: Pluralistic alignment via multi-llm collaboration.
\newblock \emph{arXiv preprint arXiv:2406.15951}, 2024.

\bibitem[Fu et~al.(2025)Fu, Zhao, Yao, Wang, Han, and Xiao]{fu2025reward}
Jiayi Fu, Xuandong Zhao, Chengyuan Yao, Heng Wang, Qi~Han, and Yanghua Xiao.
\newblock Reward shaping to mitigate reward hacking in rlhf.
\newblock \emph{arXiv preprint arXiv:2502.18770}, 2025.

\bibitem[Fu et~al.(2023)Fu, Niu, and Maher]{ZHE2023DLforSerendipity}
Zhe Fu, Xi~Niu, and Mary~Lou Maher.
\newblock Deep learning models for serendipity recommendations: A survey and new perspectives.
\newblock \emph{ACM Comput. Surv.}, 56\penalty0 (1), August 2023.
\newblock ISSN 0360-0300.
\newblock \doi{10.1145/3605145}.
\newblock URL \url{https://doi.org/10.1145/3605145}.

\bibitem[Gao et~al.(2021)Gao, Lei, He, De~Rijke, and Chua]{gao2021advances}
Chongming Gao, Wenqiang Lei, Xiangnan He, Maarten De~Rijke, and Tat-Seng Chua.
\newblock Advances and challenges in conversational recommender systems: A survey.
\newblock \emph{AI open}, 2:\penalty0 100--126, 2021.

\bibitem[Gao et~al.(2023)Gao, Gao, Dong, Tarokh, Chi, and Pajic]{gao2023off}
Qitong Gao, Ge~Gao, Juncheng Dong, Vahid Tarokh, Min Chi, and Miroslav Pajic.
\newblock Off-policy evaluation for human feedback.
\newblock \emph{Advances in Neural Information Processing Systems}, 36:\penalty0 9065--9091, 2023.

\bibitem[Grau-Moya et~al.(2018)Grau-Moya, Leibfried, and Vrancx]{grau2018soft}
Jordi Grau-Moya, Felix Leibfried, and Peter Vrancx.
\newblock Soft q-learning with mutual-information regularization.
\newblock In \emph{International conference on learning representations}, 2018.

\bibitem[Haarnoja et~al.(2017)Haarnoja, Tang, Abbeel, and Levine]{haarnoja2017reinforcement}
Tuomas Haarnoja, Haoran Tang, Pieter Abbeel, and Sergey Levine.
\newblock Reinforcement learning with deep energy-based policies.
\newblock In \emph{International conference on machine learning}, pages 1352--1361. PMLR, 2017.

\bibitem[He et~al.(2023)He, Xie, Jha, Steck, Liang, Feng, Majumder, Kallus, and McAuley]{Zhankui2023llmcrs}
Zhankui He, Zhouhang Xie, Rahul Jha, Harald Steck, Dawen Liang, Yesu Feng, Bodhisattwa~Prasad Majumder, Nathan Kallus, and Julian~J. McAuley.
\newblock Large language models as zero-shot conversational recommenders.
\newblock In \emph{Proceedings of the 32nd {ACM} International Conference on Information and Knowledge Management, {CIKM} 2023, Birmingham, United Kingdom, October 21-25, 2023}, pages 720--730. {ACM}, 2023.

\bibitem[He-Yueya et~al.(2024)He-Yueya, Ma, Gandhi, Domingue, Brunskill, and Goodman]{heyueya2024psychometricalignmentcapturinghuman}
Joy He-Yueya, Wanjing~Anya Ma, Kanishk Gandhi, Benjamin~W. Domingue, Emma Brunskill, and Noah~D. Goodman.
\newblock Psychometric alignment: Capturing human knowledge distributions via language models, 2024.
\newblock URL \url{https://arxiv.org/abs/2407.15645}.

\bibitem[Hou et~al.(2024)Hou, Li, He, Yan, Chen, and McAuley]{Yupeng2024bridging}
Yupeng Hou, Jiacheng Li, Zhankui He, An~Yan, Xiusi Chen, and Julian~J. McAuley.
\newblock Bridging language and items for retrieval and recommendation.
\newblock \emph{CoRR}, abs/2403.03952, 2024.

\bibitem[Huang et~al.(2023{\natexlab{a}})Huang, Wang, Wang, and Yao]{huang1}
Chengkai Huang, Shoujin Wang, Xianzhi Wang, and Lina Yao.
\newblock Dual contrastive transformer for hierarchical preference modeling in sequential recommendation.
\newblock In \emph{Proceedings of the 46th International {ACM} {SIGIR} Conference on Research and Development in Information Retrieval, {SIGIR} 2023, Taipei, Taiwan, July 23-27, 2023}, pages 99--109. {ACM}, 2023{\natexlab{a}}.

\bibitem[Huang et~al.(2023{\natexlab{b}})Huang, Wang, Wang, and Yao]{huang2}
Chengkai Huang, Shoujin Wang, Xianzhi Wang, and Lina Yao.
\newblock Modeling temporal positive and negative excitation for sequential recommendation.
\newblock In \emph{Proceedings of the {ACM} Web Conference 2023, {WWW} 2023, Austin, TX, USA, 30 April 2023 - 4 May 2023}, pages 1252--1263. {ACM}, 2023{\natexlab{b}}.

\bibitem[Huang et~al.(2024)Huang, Yu, Xie, Zhang, Yao, and McAuley]{fmsurvey}
Chengkai Huang, Tong Yu, Kaige Xie, Shuai Zhang, Lina Yao, and Julian~J. McAuley.
\newblock Foundation models for recommender systems: {A} survey and new perspectives.
\newblock \emph{CoRR}, abs/2402.11143, 2024.

\bibitem[Huang et~al.(2025)Huang, Wu, Xia, Yu, Wang, Yu, Zhang, Rossi, Kveton, Zhou, McAuley, and Yao]{agentsurvey}
Chengkai Huang, Junda Wu, Yu~Xia, Zixu Yu, Ruhan Wang, Tong Yu, Ruiyi Zhang, Ryan~A. Rossi, Branislav Kveton, Dongruo Zhou, Julian~J. McAuley, and Lina Yao.
\newblock Towards agentic recommender systems in the era of multimodal large language models.
\newblock \emph{CoRR}, abs/2503.16734, 2025.

\bibitem[Hurst et~al.(2024)Hurst, Lerer, Goucher, Perelman, Ramesh, Clark, Ostrow, Welihinda, Hayes, Radford, et~al.]{hurst2024gpt}
Aaron Hurst, Adam Lerer, Adam~P Goucher, Adam Perelman, Aditya Ramesh, Aidan Clark, AJ~Ostrow, Akila Welihinda, Alan Hayes, Alec Radford, et~al.
\newblock Gpt-4o system card.
\newblock \emph{arXiv preprint arXiv:2410.21276}, 2024.

\bibitem[Imambi et~al.(2021)Imambi, Prakash, and Kanagachidambaresan]{imambi2021pytorch}
Sagar Imambi, Kolla~Bhanu Prakash, and GR~Kanagachidambaresan.
\newblock Pytorch.
\newblock \emph{Programming with TensorFlow: solution for edge computing applications}, pages 87--104, 2021.

\bibitem[Ionides(2008)]{ionides2008truncated}
Edward~L Ionides.
\newblock Truncated importance sampling.
\newblock \emph{Journal of Computational and Graphical Statistics}, 17\penalty0 (2):\penalty0 295--311, 2008.

\bibitem[Jafari~Ziarani and Ravanmehr(2021)]{ziarani2021serendipity}
Reza Jafari~Ziarani and Reza Ravanmehr.
\newblock Serendipity in recommender systems: A systematic literature review.
\newblock \emph{Journal of Computer Science and Technology}, 36\penalty0 (1):\penalty0 119--138, 2021.
\newblock \doi{10.1007/s11390-020-0135-9}.
\newblock URL \url{https://link.springer.com/article/10.1007/s11390-020-0135-9}.

\bibitem[Jang et~al.(2023{\natexlab{a}})Jang, Kim, Lin, Wang, Hessel, Zettlemoyer, Hajishirzi, Choi, and Ammanabrolu]{Joel2023personalizedsoup}
Joel Jang, Seungone Kim, Bill~Yuchen Lin, Yizhong Wang, Jack Hessel, Luke Zettlemoyer, Hannaneh Hajishirzi, Yejin Choi, and Prithviraj Ammanabrolu.
\newblock Personalized soups: Personalized large language model alignment via post-hoc parameter merging.
\newblock \emph{CoRR}, abs/2310.11564, 2023{\natexlab{a}}.

\bibitem[Jang et~al.(2023{\natexlab{b}})Jang, Kim, Lin, Wang, Hessel, Zettlemoyer, Hajishirzi, Choi, and Ammanabrolu]{jang2023personalizedsoupspersonalizedlarge}
Joel Jang, Seungone Kim, Bill~Yuchen Lin, Yizhong Wang, Jack Hessel, Luke Zettlemoyer, Hannaneh Hajishirzi, Yejin Choi, and Prithviraj Ammanabrolu.
\newblock Personalized soups: Personalized large language model alignment via post-hoc parameter merging, 2023{\natexlab{b}}.
\newblock URL \url{https://arxiv.org/abs/2310.11564}.

\bibitem[Ji et~al.(2025)Ji, Wu, Wu, Wang, Yang, Dras, and Naseem]{ji2025survey}
Miaomiao Ji, Yanqiu Wu, Zhibin Wu, Shoujin Wang, Jian Yang, Mark Dras, and Usman Naseem.
\newblock A survey on progress in llm alignment from the perspective of reward design.
\newblock \emph{arXiv preprint arXiv:2505.02666}, 2025.

\bibitem[Ji et~al.(2024)Ji, Kulkarni, Wang, and Xie]{ji2024self}
Xiang Ji, Sanjeev Kulkarni, Mengdi Wang, and Tengyang Xie.
\newblock Self-play with adversarial critic: Provable and scalable offline alignment for language models.
\newblock \emph{arXiv preprint arXiv:2406.04274}, 2024.

\bibitem[Jiang and Li(2016)]{jiang2016doubly}
Nan Jiang and Lihong Li.
\newblock Doubly robust off-policy value evaluation for reinforcement learning.
\newblock In \emph{International conference on machine learning}, pages 652--661. PMLR, 2016.

\bibitem[Kingma and Ba(2014)]{kingma2014adam}
Diederik~P Kingma and Jimmy Ba.
\newblock Adam: A method for stochastic optimization.
\newblock \emph{arXiv preprint arXiv:1412.6980}, 2014.

\bibitem[Knees et~al.(2023)Knees, Neidhardt, and Nalis]{knees2023recommender}
Peter Knees, Julia Neidhardt, and Irina Nalis.
\newblock Recommender systems: Techniques, effects, and measures toward pluralism and fairness.
\newblock In Francesco Ricci, Lior Rokach, Bracha Shapira, and Paul~B. Kantor, editors, \emph{Recommender Systems Handbook}. Springer, 2023.
\newblock \doi{10.1007/978-3-031-45304-5_27}.
\newblock URL \url{https://link.springer.com/chapter/10.1007/978-3-031-45304-5_27}.

\bibitem[Kumar et~al.(2024)Kumar, Park, Tsvetkov, Smith, and Hajishirzi]{kumar2024compo}
Sachin Kumar, Chan~Young Park, Yulia Tsvetkov, Noah~A Smith, and Hannaneh Hajishirzi.
\newblock Compo: Community preferences for language model personalization.
\newblock \emph{arXiv preprint arXiv:2410.16027}, 2024.

\bibitem[Lee et~al.(2024)Lee, Park, Kim, and Seo]{lee2024aligning}
Seongyun Lee, Sue~Hyun Park, Seungone Kim, and Minjoon Seo.
\newblock Aligning to thousands of preferences via system message generalization.
\newblock \emph{arXiv preprint arXiv:2405.17977}, 2024.

\bibitem[Li et~al.(2024{\natexlab{a}})Li, Zhang, Dong, Deik, Tang, and Liu]{li2024aligningcrowdfeedbackdistributional}
Dexun Li, Cong Zhang, Kuicai Dong, Derrick Goh~Xin Deik, Ruiming Tang, and Yong Liu.
\newblock Aligning crowd feedback via distributional preference reward modeling, 2024{\natexlab{a}}.
\newblock URL \url{https://arxiv.org/abs/2402.09764}.

\bibitem[Li et~al.(2015)Li, Galley, Brockett, Gao, and Dolan]{li2015diversity}
Jiwei Li, Michel Galley, Chris Brockett, Jianfeng Gao, and Bill Dolan.
\newblock A diversity-promoting objective function for neural conversation models.
\newblock \emph{arXiv preprint arXiv:1510.03055}, 2015.

\bibitem[Li et~al.(2024{\natexlab{b}})Li, Ji, Chen, and Wang]{li2024policy}
Zihao Li, Xiang Ji, Minshuo Chen, and Mengdi Wang.
\newblock Policy evaluation for reinforcement learning from human feedback: A sample complexity analysis.
\newblock In \emph{International Conference on Artificial Intelligence and Statistics}, pages 2737--2745. PMLR, 2024{\natexlab{b}}.

\bibitem[Liu et~al.(2024)Liu, Li, Huang, Luo, and Liu]{zhiqiang}
Zhiqiang Liu, Yuhong Li, Chengkai Huang, KunTing Luo, and Yanxia Liu.
\newblock Boosting fine-tuning via conditional online knowledge transfer.
\newblock \emph{Neural Networks}, 169:\penalty0 325--333, 2024.

\bibitem[Ouyang et~al.(2022)Ouyang, Wu, Jiang, Almeida, Wainwright, Mishkin, Zhang, Agarwal, Slama, Ray, Schulman, Hilton, Kelton, Miller, Simens, Askell, Welinder, Christiano, Leike, and Lowe]{ouyang2022traininglanguagemodelsfollow}
Long Ouyang, Jeff Wu, Xu~Jiang, Diogo Almeida, Carroll~L. Wainwright, Pamela Mishkin, Chong Zhang, Sandhini Agarwal, Katarina Slama, Alex Ray, John Schulman, Jacob Hilton, Fraser Kelton, Luke Miller, Maddie Simens, Amanda Askell, Peter Welinder, Paul Christiano, Jan Leike, and Ryan Lowe.
\newblock Training language models to follow instructions with human feedback, 2022.
\newblock URL \url{https://arxiv.org/abs/2203.02155}.

\bibitem[Pan et~al.(2025)Pan, Cao, Huang, Zhang, and Yao]{bias2}
Le~Pan, Yuanjiang Cao, Chengkai Huang, Wenjie Zhang, and Lina Yao.
\newblock Counterfactual inference for eliminating sentiment bias in recommender systems.
\newblock \emph{CoRR}, abs/2505.03655, 2025.

\bibitem[Peng et~al.(2023)Peng, Li, He, Galley, and Gao]{AlpacaGPT42023}
Baolin Peng, Chunyuan Li, Pengcheng He, Michel Galley, and Jianfeng Gao.
\newblock Instruction tuning with {GPT-4}.
\newblock \emph{CoRR}, abs/2304.03277, 2023.

\bibitem[Plackett(1975)]{plackett1975analysis}
Robin~L Plackett.
\newblock The analysis of permutations.
\newblock \emph{Journal of the Royal Statistical Society Series C: Applied Statistics}, 24\penalty0 (2):\penalty0 193--202, 1975.

\bibitem[Poddar et~al.(2024)Poddar, Wan, Ivison, Gupta, and Jaques]{poddar2024personalizing}
Sriyash Poddar, Yanming Wan, Hamish Ivison, Abhishek Gupta, and Natasha Jaques.
\newblock Personalizing reinforcement learning from human feedback with variational preference learning.
\newblock \emph{arXiv preprint arXiv:2408.10075}, 2024.

\bibitem[Radford et~al.(2019)Radford, Wu, Child, Luan, Amodei, Sutskever, et~al.]{radford2019language}
Alec Radford, Jeffrey Wu, Rewon Child, David Luan, Dario Amodei, Ilya Sutskever, et~al.
\newblock Language models are unsupervised multitask learners.
\newblock \emph{OpenAI blog}, 1\penalty0 (8):\penalty0 9, 2019.

\bibitem[Rafailov et~al.(2023)Rafailov, Sharma, Mitchell, Manning, Ermon, and Finn]{rafailov2023direct}
Rafael Rafailov, Archit Sharma, Eric Mitchell, Christopher~D Manning, Stefano Ermon, and Chelsea Finn.
\newblock Direct preference optimization: Your language model is secretly a reward model.
\newblock \emph{Advances in Neural Information Processing Systems}, 36:\penalty0 53728--53741, 2023.

\bibitem[Reimers and Gurevych(2019)]{reimers2019sentence}
Nils Reimers and Iryna Gurevych.
\newblock Sentence-bert: Sentence embeddings using siamese bert-networks.
\newblock \emph{arXiv preprint arXiv:1908.10084}, 2019.

\bibitem[Santurkar et~al.(2023)Santurkar, Durmus, Ladhak, Lee, Liang, and Hashimoto]{Shibani2023whose}
Shibani Santurkar, Esin Durmus, Faisal Ladhak, Cinoo Lee, Percy Liang, and Tatsunori Hashimoto.
\newblock Whose opinions do language models reflect?
\newblock In \emph{Proceedings of the 40th International Conference on Machine Learning}, ICML'23. JMLR.org, 2023.

\bibitem[Sch{\"u}tze et~al.(2008)Sch{\"u}tze, Manning, and Raghavan]{schutze2008introduction}
Hinrich Sch{\"u}tze, Christopher~D Manning, and Prabhakar Raghavan.
\newblock \emph{Introduction to information retrieval}, volume~39.
\newblock Cambridge University Press Cambridge, 2008.

\bibitem[Siththaranjan et~al.(2024)Siththaranjan, Laidlaw, and Hadfield-Menell]{siththaranjan2024distributional}
Anand Siththaranjan, Cassidy Laidlaw, and Dylan Hadfield-Menell.
\newblock Distributional preference learning: Understanding and accounting for hidden context in {RLHF}.
\newblock In \emph{The Twelfth International Conference on Learning Representations}, 2024.
\newblock URL \url{https://openreview.net/forum?id=0tWTxYYPnW}.

\bibitem[Sorensen et~al.(2024)Sorensen, Moore, Fisher, Gordon, Mireshghallah, Rytting, Ye, Jiang, Lu, Dziri, Althoff, and Choi]{Sorensen2024RoadmapPluralistic}
Taylor Sorensen, Jared Moore, Jillian Fisher, Mitchell Gordon, Niloofar Mireshghallah, Christopher~Michael Rytting, Andre Ye, Liwei Jiang, Ximing Lu, Nouha Dziri, Tim Althoff, and Yejin Choi.
\newblock Position: a roadmap to pluralistic alignment.
\newblock ICML'24. JMLR.org, 2024.

\bibitem[Stiennon et~al.(2022)Stiennon, Ouyang, Wu, Ziegler, Lowe, Voss, Radford, Amodei, and Christiano]{stiennon2022learningsummarizehumanfeedback}
Nisan Stiennon, Long Ouyang, Jeff Wu, Daniel~M. Ziegler, Ryan Lowe, Chelsea Voss, Alec Radford, Dario Amodei, and Paul Christiano.
\newblock Learning to summarize from human feedback, 2022.
\newblock URL \url{https://arxiv.org/abs/2009.01325}.

\bibitem[Vermeulen(2022)]{vermeulen2022nudge}
Judith Vermeulen.
\newblock To nudge or not to nudge: News recommendation as a tool to achieve online media pluralism.
\newblock \emph{Journal of Media Law}, X\penalty0 (X):\penalty0 1--18, 2022.
\newblock \doi{10.1080/21670811.2021.2021804}.
\newblock URL \url{https://doi.org/10.1080/21670811.2021.2021804}.

\bibitem[Wang et~al.(2024)Wang, Yang, Zhu, Yang, Cohen, Li, and Tian]{wang2024learningpersonalizedalignmentevaluating}
Danqing Wang, Kevin Yang, Hanlin Zhu, Xiaomeng Yang, Andrew Cohen, Lei Li, and Yuandong Tian.
\newblock Learning personalized alignment for evaluating open-ended text generation, 2024.
\newblock URL \url{https://arxiv.org/abs/2310.03304}.

\bibitem[Welinder et~al.(2010)Welinder, Branson, Perona, and Belongie]{welinder2010multidimensional}
Peter Welinder, Steve Branson, Pietro Perona, and Serge Belongie.
\newblock The multidimensional wisdom of crowds.
\newblock \emph{Advances in neural information processing systems}, 23, 2010.

\bibitem[Wu et~al.(2024)Wu, Li, Wang, Xia, Xiong, Wang, Yu, Chen, Kveton, Yao, et~al.]{wu2024ocean}
Junda Wu, Xintong Li, Ruoyu Wang, Yu~Xia, Yuxin Xiong, Jianing Wang, Tong Yu, Xiang Chen, Branislav Kveton, Lina Yao, et~al.
\newblock Ocean: Offline chain-of-thought evaluation and alignment in large language models.
\newblock \emph{arXiv preprint arXiv:2410.23703}, 2024.

\bibitem[Xie et~al.(2025)Xie, Wu, Shen, Xia, Li, Chang, Rossi, Kumar, Majumder, Shang, Ammanabrolu, and McAuley]{xie2025surveypersonalizedpluralisticpreference}
Zhouhang Xie, Junda Wu, Yiran Shen, Yu~Xia, Xintong Li, Aaron Chang, Ryan Rossi, Sachin Kumar, Bodhisattwa~Prasad Majumder, Jingbo Shang, Prithviraj Ammanabrolu, and Julian McAuley.
\newblock A survey on personalized and pluralistic preference alignment in large language models, 2025.
\newblock URL \url{https://arxiv.org/abs/2504.07070}.

\bibitem[Yang et~al.(2024)Yang, Yang, Zhang, Hui, Zheng, Yu, Li, Liu, Huang, Wei, Lin, Yang, Tu, Zhang, Yang, Yang, Zhou, Lin, Dang, Lu, Bao, Yang, Yu, Li, Xue, Zhang, Zhu, Men, Lin, Li, Xia, Ren, Ren, Fan, Su, Zhang, Wan, Liu, Cui, Zhang, and Qiu]{qwen2.5}
An~Yang, Baosong Yang, Beichen Zhang, Binyuan Hui, Bo~Zheng, Bowen Yu, Chengyuan Li, Dayiheng Liu, Fei Huang, Haoran Wei, Huan Lin, Jian Yang, Jianhong Tu, Jianwei Zhang, Jianxin Yang, Jiaxi Yang, Jingren Zhou, Junyang Lin, Kai Dang, Keming Lu, Keqin Bao, Kexin Yang, Le~Yu, Mei Li, Mingfeng Xue, Pei Zhang, Qin Zhu, Rui Men, Runji Lin, Tianhao Li, Tingyu Xia, Xingzhang Ren, Xuancheng Ren, Yang Fan, Yang Su, Yichang Zhang, Yu~Wan, Yuqiong Liu, Zeyu Cui, Zhenru Zhang, and Zihan Qiu.
\newblock Qwen2.5 technical report.
\newblock \emph{arXiv preprint arXiv:2412.15115}, 2024.

\bibitem[Yao et~al.(2025)Yao, Cai, Chuang, Yang, Jiang, Yang, and Hu]{yao2025no}
Binwei Yao, Zefan Cai, Yun-Shiuan Chuang, Shanglin Yang, Ming Jiang, Diyi Yang, and Junjie Hu.
\newblock No preference left behind: Group distributional preference optimization.
\newblock In \emph{The Thirteenth International Conference on Learning Representations}, 2025.
\newblock URL \url{https://openreview.net/forum?id=bgpNJBD6Va}.

\bibitem[Yin and Zhao(2024)]{yin2024diversity}
Kexin Yin and Junqi Zhao.
\newblock Diversity and serendipity preference-aware recommender system.
\newblock \emph{Journal of Computer Engineering \& Computer Technology}, 2024.
\newblock \doi{10.47852/bonviewJCCE42023272}.
\newblock URL \url{https://doi.org/10.47852/bonviewJCCE42023272}.

\bibitem[Yin et~al.(2024)Yin, Wang, Xie, Chen, and Zhou]{yin2024self}
Yueqin Yin, Zhendong Wang, Yujia Xie, Weizhu Chen, and Mingyuan Zhou.
\newblock Self-augmented preference optimization: Off-policy paradigms for language model alignment.
\newblock \emph{arXiv preprint arXiv:2405.20830}, 2024.

\bibitem[Zha et~al.(2023)Zha, Yang, Li, and Hu]{zha2023alignscore}
Yuheng Zha, Yichi Yang, Ruichen Li, and Zhiting Hu.
\newblock Alignscore: Evaluating factual consistency with a unified alignment function.
\newblock \emph{arXiv preprint arXiv:2305.16739}, 2023.

\bibitem[Zhang et~al.(2025)Zhang, Chong, Jiang, Tang, Gao, Tang, and Li]{zhang2025aligning}
Chen Zhang, Dading Chong, Feng Jiang, Chengguang Tang, Anningzhe Gao, Guohua Tang, and Haizhou Li.
\newblock Aligning language models using follow-up likelihood as reward signal.
\newblock In \emph{Proceedings of the AAAI Conference on Artificial Intelligence}, volume~39, pages 25832--25841, 2025.

\bibitem[Zhang et~al.(2024{\natexlab{a}})Zhang, Wang, Hwang, Dong, Delalleau, Choi, Choi, Ren, and Pyatkin]{zhang2024divergingpreferencesannotatorsdisagree}
Michael~JQ Zhang, Zhilin Wang, Jena~D. Hwang, Yi~Dong, Olivier Delalleau, Yejin Choi, Eunsol Choi, Xiang Ren, and Valentina Pyatkin.
\newblock Diverging preferences: When do annotators disagree and do models know?, 2024{\natexlab{a}}.
\newblock URL \url{https://arxiv.org/abs/2410.14632}.

\bibitem[Zhang et~al.(2024{\natexlab{b}})Zhang, Peng, and Bollegala]{zhang2024improving}
Tianhui Zhang, Bei Peng, and Danushka Bollegala.
\newblock Improving diversity of commonsense generation by large language models via in-context learning.
\newblock \emph{arXiv preprint arXiv:2404.16807}, 2024{\natexlab{b}}.

\bibitem[Zhao et~al.(2023{\natexlab{a}})Zhao, Dang, and Grover]{zhao2023group}
Siyan Zhao, John Dang, and Aditya Grover.
\newblock Group preference optimization: Few-shot alignment of large language models.
\newblock \emph{arXiv preprint arXiv:2310.11523}, 2023{\natexlab{a}}.

\bibitem[Zhao et~al.(2023{\natexlab{b}})Zhao, Wang, Liu, Cheng, Aggarwal, and Derr]{zhao2023fairness}
Yuying Zhao, Yu~Wang, Yunchao Liu, Xueqi Cheng, Charu~C. Aggarwal, and Tyler Derr.
\newblock Fairness and diversity in recommender systems: A survey.
\newblock \emph{ACM Transactions on Intelligent Systems and Technology}, 2023{\natexlab{b}}.
\newblock URL \url{https://arxiv.org/abs/2307.04644}.
\newblock preprint arXiv:2307.04644.

\bibitem[Zheng et~al.(2015)Zheng, Chan, and Ip]{zheng2015unexpectedness}
Qianru Zheng, Chi-Kong Chan, and Horace~HS Ip.
\newblock An unexpectedness-augmented utility model for making serendipitous recommendation.
\newblock In \emph{Advances in Data Mining: Applications and Theoretical Aspects: 15th Industrial Conference, ICDM 2015, Hamburg, Germany, July 11-24, 2015, Proceedings 15}, pages 216--230. Springer, 2015.

\bibitem[Zhou et~al.(2024)Zhou, Agrawal, Zhang, Indurthi, Zhao, Song, Xu, and Zhu]{zhou2024wpo}
Wenxuan Zhou, Ravi Agrawal, Shujian Zhang, Sathish~Reddy Indurthi, Sanqiang Zhao, Kaiqiang Song, Silei Xu, and Chenguang Zhu.
\newblock Wpo: Enhancing rlhf with weighted preference optimization.
\newblock \emph{arXiv preprint arXiv:2406.11827}, 2024.

\bibitem[Zhu et~al.(2018)Zhu, Lu, Zheng, Guo, Zhang, Wang, and Yu]{zhu2018texygen}
Yaoming Zhu, Sidi Lu, Lei Zheng, Jiaxian Guo, Weinan Zhang, Jun Wang, and Yong Yu.
\newblock Texygen: A benchmarking platform for text generation models.
\newblock In \emph{The 41st international ACM SIGIR conference on research \& development in information retrieval}, pages 1097--1100, 2018.

\end{thebibliography}

\section{Decomposable Lower Bound via Maximum Entropy Principle}\label{app:entropy}

\begin{thm} \label{app:maxent-lower-bound}
We aim to evaluate the expected reward under the target policy $\pi$ using samples from $\pi_0$, which gives:
\begin{equation}
    \mathbb{E}_{S_t \sim \pi_0} \left[ \frac{\pi(S_t \mid x_t)}{\pi_0(S_t \mid x_t)} R(x_t, S_t) \right].
\end{equation}

Assuming the responses $a_{t,i}$ are drawn independently under $\pi_0(\cdot \mid x_t)$, we approximate:
\begin{equation}
    \frac{\pi(S_t \mid x_t)}{\pi_0(S_t \mid x_t)} \approx \prod_{i=1}^K \frac{\pi(a_{t,i} \mid x_t)}{\pi_0(a_{t,i} \mid x_t)}.
\end{equation}

Now define for each individual action the reward:
\begin{equation}
    r(x_t, a_{t,i}) := \eta_{t,i}(x_t, a_{t,i}) - \log \pi(a_{t,i} \mid x_t).
\end{equation}

Then we use Jensen's inequality on the convex function \( f(z) = z \cdot R(x_t, S_t) \), applied to the product distribution:
\begin{align}
    \mathbb{E}_{S_t \sim \pi_0} \left[ \prod_{i=1}^K \frac{\pi(a_{t,i} \mid x_t)}{\pi_0(a_{t,i} \mid x_t)} R(x_t, S_t) \right]
    &\geq \sum_{i=1}^K \mathbb{E}_{a_{t,i} \sim \pi_0} \left[ \frac{\pi(a_{t,i} \mid x_t)}{\pi_0(a_{t,i} \mid x_t)} r(x_t, a_{t,i}) \right] \\
    &= \sum_{i=1}^K \mathbb{E}_{a_{t,i} \sim \pi_0} \left[ \frac{\pi(a_{t,i} \mid x_t)}{\pi_0(a_{t,i} \mid x_t)} \left( \eta_{t,i}(x_t, a_{t,i}) - \log \pi(a_{t,i} \mid x_t) \right) \right].
\end{align}

\noindent\textbf{Jensen’s inequality step.}
Let
\[
A_t \;=\; 
\Bigl(\prod_{i=1}^K w_{t,i}\Bigr)\Bigl(\sum_{i=1}^K r(x_t,a_{t,i})\Bigr).
\]
Since \(\log\) is concave,
\[
\log\bigl(\mathbb{E}_{S_t\sim\pi_0}[A_t]\bigr)
\;\ge\;
\mathbb{E}_{S_t\sim\pi_0}\bigl[\log A_t\bigr].
\]
Exponentiating both sides gives
\[
\mathbb{E}_{S_t\sim\pi_0}[A_t]
\;\ge\;
\exp\Bigl(\mathbb{E}_{S_t\sim\pi_0}[\log A_t]\Bigr).
\]
But
\[
\log A_t
= \sum_{i=1}^K \log w_{t,i}
\;+\;\log\!\Bigl(\sum_{i=1}^K r(x_t,a_{t,i})\Bigr),
\]
and an application of Jensen’s inequality once more (this time to the concave \(\log\) on the sum inside) shows
\[
\log\!\Bigl(\sum_{i=1}^K r(x_t,a_{t,i})\Bigr)
\;\ge\;
\sum_{i=1}^K
\frac{r(x_t,a_{t,i})}{\sum_j r(x_t,a_{t,j})}
\;\log\bigl(r(x_t,a_{t,i})\bigr).
\]
Combining these facts and using the nonnegativity of all weights and rewards yields
\[
\mathbb{E}_{S_t}\Bigl[A_t\Bigr]
\;\ge\;
\sum_{i=1}^K
\mathbb{E}_{a_{t,i}\sim\pi_0}\Bigl[w_{t,i}\,r(x_t,a_{t,i})\Bigr]
=
\sum_{i=1}^K
\mathbb{E}_{a_{t,i}\sim\pi_0}\!\Bigl[\tfrac{\pi(a_{t,i}\!\mid x_t)}{\pi_0(a_{t,i}\!\mid x_t)}
\bigl(\eta_{t,i}-\log\pi(a_{t,i}\!\mid x_t)\bigr)\Bigr].
\]
Averaging over \(t=1,\dots,n\) completes the proof of the bound \eqref{eq:entropy-lowerbound}.

Averaging over all $t = 1, \ldots, n$ completes the result:
\begin{equation}
    \hat{V}_\mathrm{POPE}(\pi)
    \geq \frac{1}{n} \sum_{t=1}^n \sum_{i=1}^K \frac{\pi(a_{t,i} \mid x_t)}{\pi_0(a_{t,i} \mid x_t)} \left[ \eta_{t,i}(x_t, a_{t,i}) - \log \pi(a_{t,i} \mid x_t) \right].
\end{equation}

\end{thm}

\section{Case Study}

\textbf{Please Write helpful reviews for this product: \textbf{This box set includes the complete sixth season of SpongeBob Squarepants.}}

\begin{tcolorbox}[colback=blue!5,colframe=blue!40,title={Phi-3.5 Base Model Response}]
The set includes all 25 episodes.

The box set was released on December 17, 2012.

References

External links

SpongeBob SquarePants (2012 album)  
SpongeBob SquarePants (2012 video game)  
SpongeBob SquarePants (2012 video game)  
SpongeBob SquarePants (2012 album)
\end{tcolorbox}

\begin{tcolorbox}[colback=purple!5,colframe=purple!40,title={SFT Response}]
It has 14 episodes and comes in a very nice looking box. The box also has a nice looking cover and an insert with a mini-game. The box is quite big, so it is probably best to put it on a shelf. It is not a cheap box set, but it is worth it. The box set is in perfect condition and the episodes are in great shape. I’ve watched all of the episodes and the box set is perfect for me.
\end{tcolorbox}

\begin{tcolorbox}[colback=green!5,colframe=green!40,title={DPO Response}]
The sixth season began airing in the United States on May 24, 2006.

In the United States, it aired on Nickelodeon, and was released on DVD on May 13, 2009.

In the United Kingdom, it aired on Nickelodeon and Nicktoons, and was released on DVD on October 31, 2009.

\end{tcolorbox}

\begin{tcolorbox}[colback=orange!5,colframe=orange!40,title={\texttt{POPE} Response}]
It features 26 episodes, including the original pilot “Help Wanted” and the episode “SpongeBob SquarePants: The Movie – Sponge Out of Water,” which was released in 2016. The collection also includes special features such as commentaries, deleted scenes, and behind-the-scenes footage. This set is perfect for fans looking to revisit the beloved adventures of SpongeBob and his friends.
\end{tcolorbox}

These results are based on the Phi-3.5 model evaluated on the Amazon Movie Reviews dataset, comparing Phi-Base, SFT, DPO, and our \texttt{POPE} models. The case study clearly illustrates how \texttt{POPE} substantially improves the quality and breadth of generated reviews. Whereas the base and SFT models either omit key details or focus narrowly on formatting and packaging, and DPO reproduces only factual release information, our \texttt{POPE}-aligned model synthesizes a more complete, fan-centered review that balances episode highlights (“Help Wanted,” “Sponge Out of Water”), production facts (special features, deleted scenes), and a recommendation voice. By optimizing for the full distribution of human responses rather than a single majority view, \texttt{POPE} enables the LLM to produce reviews that are simultaneously informative, engaging, and reflective of diverse user preferences.

\section{Evaluation Metrics Details}\label{app:metrics}


Alongside the off-policy evaluation framework presented in Section \ref{sec:PPE}, we employ both established and newly proposed, task-specific metrics to comprehensively assess our models’ capabilities:
\subsection{RQ1 Evaluation Metrics}
\begin{enumerate}[label=(\roman*)]
  \item \textbf{PL-Score (higher is better)}: To measure how well LLM’s response rankings match human Plackett–Luce preference judgments. Let $M$ be the number of ground‐truth responses $\{\mathrm{gt}_k\}_{k=1}^M$ with up‐votes $u_k$, and $N$ the number of generated candidates $\{\mathrm{gen}_j\}_{j=1}^N$. We define
  \begin{align}
      \theta_k = \frac{u_k}{\sum_{\ell=1}^M u_\ell},
    \quad
    S_{j,k} = \cos\bigl(\mathbf{e}(\mathrm{gen}_j),\,\mathbf{e}(\mathrm{gt}_k)\bigr).
  \end{align}
  Then
  \begin{align}
    \mathrm{PL\text{-}Score}
    = \frac{1}{N}\sum_{j=1}^N\sum_{k=1}^M \theta_k\,S_{j,k}.
  \end{align}  
  
  \item \textbf{Pluralistic Coverage (higher is better)}: We define the Pluralistic Coverage as the fraction of distinct reference responses retrieved among the top‐$k$ outputs above a similarity threshold $\delta$. Formally,
  \begin{align}
      \mathrm{Coverage}
    = \frac{1}{M}\sum_{k=1}^M
      \mathbb{I}\bigl(\max_{1\le j\le N}S_{j,k}>\delta\bigr).
  \end{align}
  In this paper, we adopt $\delta$ as 0.8.
  \item \textbf{Distributional Alignment (higher is better)}: normalized entropy of the implicit distribution over ground‐truth responses, measuring diversity of the model’s slate. Let
  \begin{align}
    p_k = \frac{\exp\bigl(\sum_{j=1}^N S_{j,k}/\tau\bigr)}{\sum_{\ell=1}^M \exp\bigl(\sum_{j=1}^N S_{j,\ell}/\tau\bigr)}
    \quad(\tau\text{ is a temperature}).
  \end{align}
  Then
    \begin{align}
    \mathrm{Distributional \ Alignment}
    = -\frac{1}{\log M}\sum_{k=1}^M p_k\log p_k
  \end{align}
  We set $\tau$ as 0.5 in this paper. It is worth noting that distributional alignment quantifies the normalized Shannon entropy of the model’s induced preference distribution over human references, akin to how Jensen–Shannon divergence (JSD) \cite{zhao2023group} measures distributional differences in prior work.
  \item  \textbf{Diversity \cite{fang2022investigating,zhang2024improving} (higher is better)}: measures the average pairwise diversity of the generated candidates. 
  Let
  \begin{align}
      \mathrm{Diversity}
    = 1 - \frac{1}{C\binom{N}{2}}\sum_{1\le i<j\le N}
      \cos\bigl(\mathbf{e}(\mathrm{gen}_i),\,\mathbf{e}(\mathrm{gen}_j)\bigr).
  \end{align}
\end{enumerate}
\subsection{RQ2 and RQ3 Evaluation Metrics}
\begin{enumerate}[label=(\roman*)]
  \item \textbf{Helpfulness score (higher is better)}: for each model response $i$, let its human replies be $\{\mathrm{reply}_{i,k}\}_{k=1}^{K_i}$ with raw up‐votes $v_{i,k}$. After min–max scaling to $[0,10]$ to obtain $v'_{i,k}$ and normalizing to weights $w_{i,k}=v'_{i,k}/\sum_j v'_{i,j}$, we define
  \begin{align}
    \mathrm{Helpfulness}_i
    = \sum_{k=1}^{K_i} w_{i,k}\,\cos\bigl(\mathbf{e}(\mathrm{resp}_i),\,\mathbf{e}(\mathrm{reply}_{i,k})\bigr).
  \end{align}
  \item \textbf{Relevance score \cite{schutze2008introduction} (higher is better)}: cosine similarity between question (or query) and response embeddings:
  \begin{align}
    \mathrm{Relevance}_i
    = \cos\bigl(\mathbf{e}(\mathrm{que}_i),\,\mathbf{e}(\mathrm{resp}_i)\bigr)
  \end{align}
  Embeddings are computed via SentenceBert \cite{reimers2019sentence}.
  \item \textbf{Distinct-N  \cite{li2015diversity} (higher is better)}: measures lexical diversity of the generated responses. 
\end{enumerate}
\subsection{RQ4 Evaluation Metrics}
\textbf{Coverage (higher is better)}: fraction of unique human reference responses covered by the model’s top‐$k$ outputs (same as Coverage above, see RQ1).
\subsection{RQ5 Evaluation Metrics}
\begin{enumerate}[label=(\roman*)]
  \item \textbf{Self-BLEU (lower is better)}: average BLEU score of each generated response against all other responses, using smoothing \cite{zhu2018texygen}.
  \item \textbf{AlignScore (higher is better)}: measures semantic faithfulness of each response given context, implemented via AlignScore \cite{zha2023alignscore}.
\end{enumerate}

Aligned with our two reward components, the collaborative‐utility metrics are PL-Score, Helpfulness, Relevance, and AlignScore, while the diversity metrics are Pluralistic Coverage, Distributional Alignment, Diversity, Distinct-N, and Self-BLEU.

\section{Synthesis of Alpaca-GPT4 Toy Dataset for RQ1 Experiment}\label{toy_dataset}

We randomly sampled 500 ‘description’ queries (i.e. prompts that ask the user to describe something) from the Alpaca-GPT4 dataset. For each query, we then instructed GPT-4o to generate five stylistically distinct responses, which serve as proxy ground truth for our pluralistic evaluation. The prompt template is shown below.

\begin{tcolorbox}[title=Prompt Template,colback=gray!5,colframe=black]
\begin{verbatim}
PROMPT_TEMPLATE = """
You are a helpful assistant. Given the user’s question, 
generate five distinct responses,
each in one of the following styles:

1. Formal Academic
2. Casual Conversational
3. Technical Bullet-Point Overview
4. Metaphorical/Poetic Description
5. Humorous Anecdotal Narrative

Return only a JSON object with keys "reply_1" through "reply_5". 
Example:
{
  "reply_1": "...Formal Academic response...",
  "reply_2": "...Casual Conversational response...",
  "reply_3": "...Technical Bullet-Point response...",
  "reply_4": "...Metaphorical/Poetic response...",
  "reply_5": "...Humorous Anecdotal response..."
}
User Question: "{instruction}"
"""
\end{verbatim}
\end{tcolorbox}

These five outputs, each reflecting a different register or tone, form our synthesized “pluralistic” set of human references for RQ1. 

\section{LLM-as-Judge Win Rate per Model Comparison}\label{sec:evaluation_metrics}

To further evaluate the effect of \texttt{POPE}, we adopt GPT-4o as an automatic judge to perform pairwise comparisons between the model outputs. Specifically, for each user query, we present GPT-4o with two sets of 50 responses generated by different models (e.g., \texttt{POPE} vs DPO), along with three human-written ground-truth responses. The model is prompted to judge which set of responses is more helpful and diverse, considering pluralistic user preferences.

\begin{figure}[ht]
    \centering
    \includegraphics[width=0.5\linewidth]{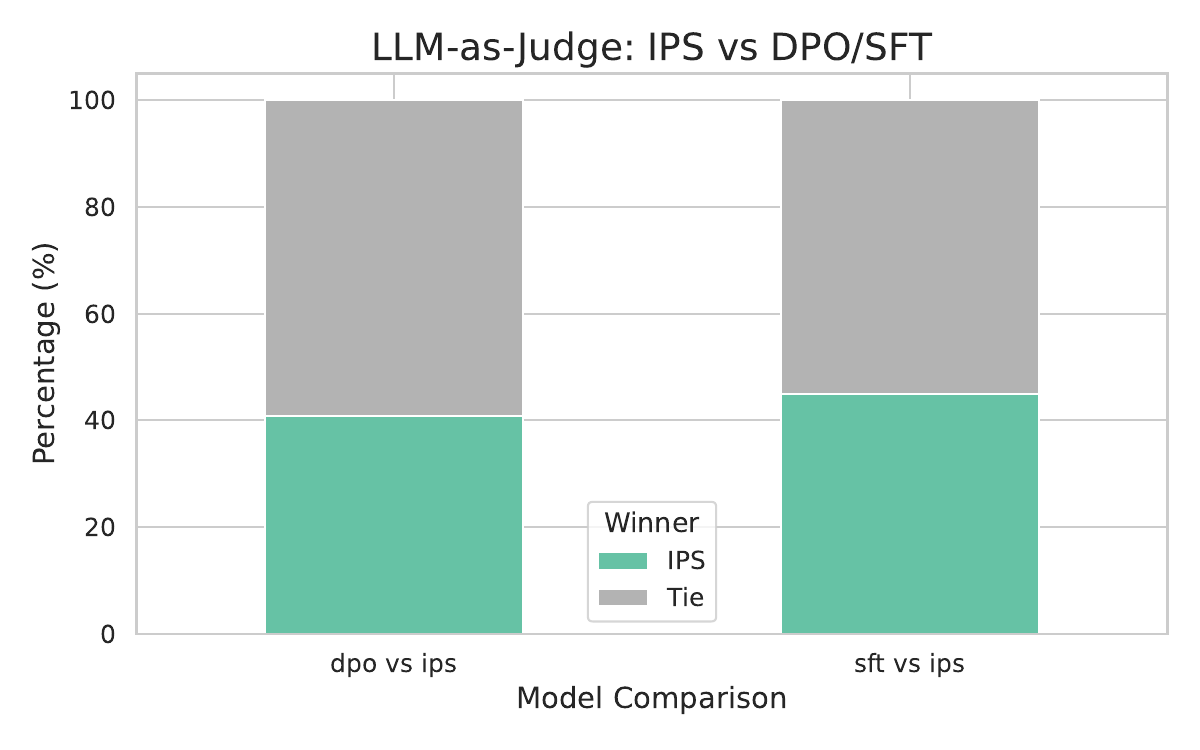}
    \caption{Win rate distribution from GPT-4o judgments comparing \texttt{POPE} with DPO and SFT on GPT4-Alpaca dataset. \texttt{POPE} is preferred more frequently, indicating better alignment with pluralistic user preferences.}
    \label{fig:enter-label}
\end{figure}

Figure~\ref{fig:enter-label} shows the win rate distribution of \texttt{POPE} against other baseline models. We observe that \texttt{POPE} consistently outperforms both DPO and SFT in terms of the number of judged wins. The relatively high proportion of ties also suggests that, while the overall quality of baseline models is strong, \texttt{POPE} introduces a broader and more pluralistic range of responses that are preferred by the LLM judge.

\section{Implementation Details}\label{app:imp_details}

We implement our model and all baseline methods using PyTorch \cite{imambi2021pytorch} and conduct experiments on four NVIDIA RTX A6000 GPUs, each equipped with 48 GB of memory. All hyperparameters of baselines are carefully tuned to ensure a fair comparison. Specifically, we train using a learning rate of 1e-5, a batch size of 4, and optimize with AdamW \cite{kingma2014adam} . 




\end{document}